\documentclass[10pt,twocolumn,letterpaper]{article}

\usepackage[pagenumbers]{cvpr} 

\usepackage[accsupp]{axessibility} 

\usepackage{graphicx}
\usepackage{amsmath}
\usepackage{amssymb}
\usepackage{booktabs}
\usepackage{diagbox}
\usepackage{ulem}
\usepackage{overpic}
\usepackage[T1]{fontenc}
\usepackage{mathtools}  
\usepackage{authblk}
\usepackage[toc,page]{appendix}

\usepackage[pagebackref,breaklinks,colorlinks]{hyperref}

\usepackage[capitalize]{cleveref}
\crefname{section}{Sec.}{Secs.}
\Crefname{section}{Section}{Sections}
\Crefname{table}{Table}{Tables}
\crefname{table}{Tab.}{Tabs.}

\def\cvprPaperID{5928} 
\def\confName{CVPR}
\def\confYear{2022}

\begin{document}

\title{Self-Supervised Image Representation Learning with Geometric Set Consistency}



\makeatletter 
\renewcommand\AB@affilsepx{\quad \protect\Affilfont} 
\makeatother

\author[1,2]{Nenglun Chen\thanks{This work was done when Nenglun Chen was an intern at Microsoft Research Asia.}}
\author[2]{Lei Chu}
\author[2]{Hao Pan}
\author[2]{Yan Lu}
\author[3]{Wenping Wang}

\affil[1]{The University of Hong Kong}
\makeatletter 
\renewcommand\AB@affilsepx{\\ \protect\Affilfont} 
\makeatother
\affil[2]{Microsoft Research Asia}
\makeatletter 
\renewcommand\AB@affilsepx{\quad \protect\Affilfont} 
\makeatother
\affil[3]{Texas A\&M University}

\maketitle

\begin{abstract}
 We propose a method for self-supervised image representation learning under the guidance of 3D geometric consistency. Our intuition is that 3D geometric consistency priors such as smooth regions and surface discontinuities may imply consistent semantics or object boundaries, and can act as strong cues to guide the learning of 2D image representations without semantic labels. Specifically, we introduce 3D geometric consistency into a contrastive learning framework to enforce the feature consistency within image views. We propose to use geometric consistency sets as constraints and adapt the InfoNCE loss accordingly. We show that our learned image representations are general. By fine-tuning our pre-trained representations for various 2D image-based downstream tasks, including semantic segmentation, object detection, and instance segmentation on real-world indoor scene datasets, we achieve superior performance compared with state-of-the-art methods.
 
\end{abstract}

\section{Introduction}

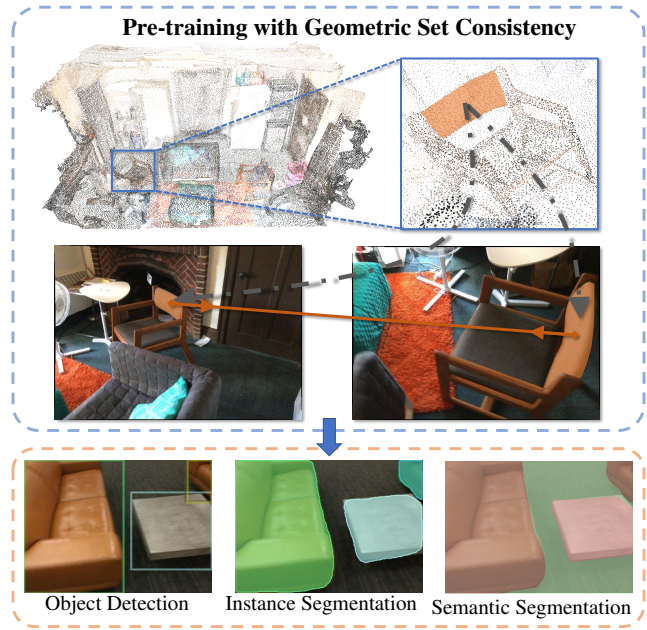
\begin{figure}
\centering
\begin{overpic}[width=0.49\textwidth]{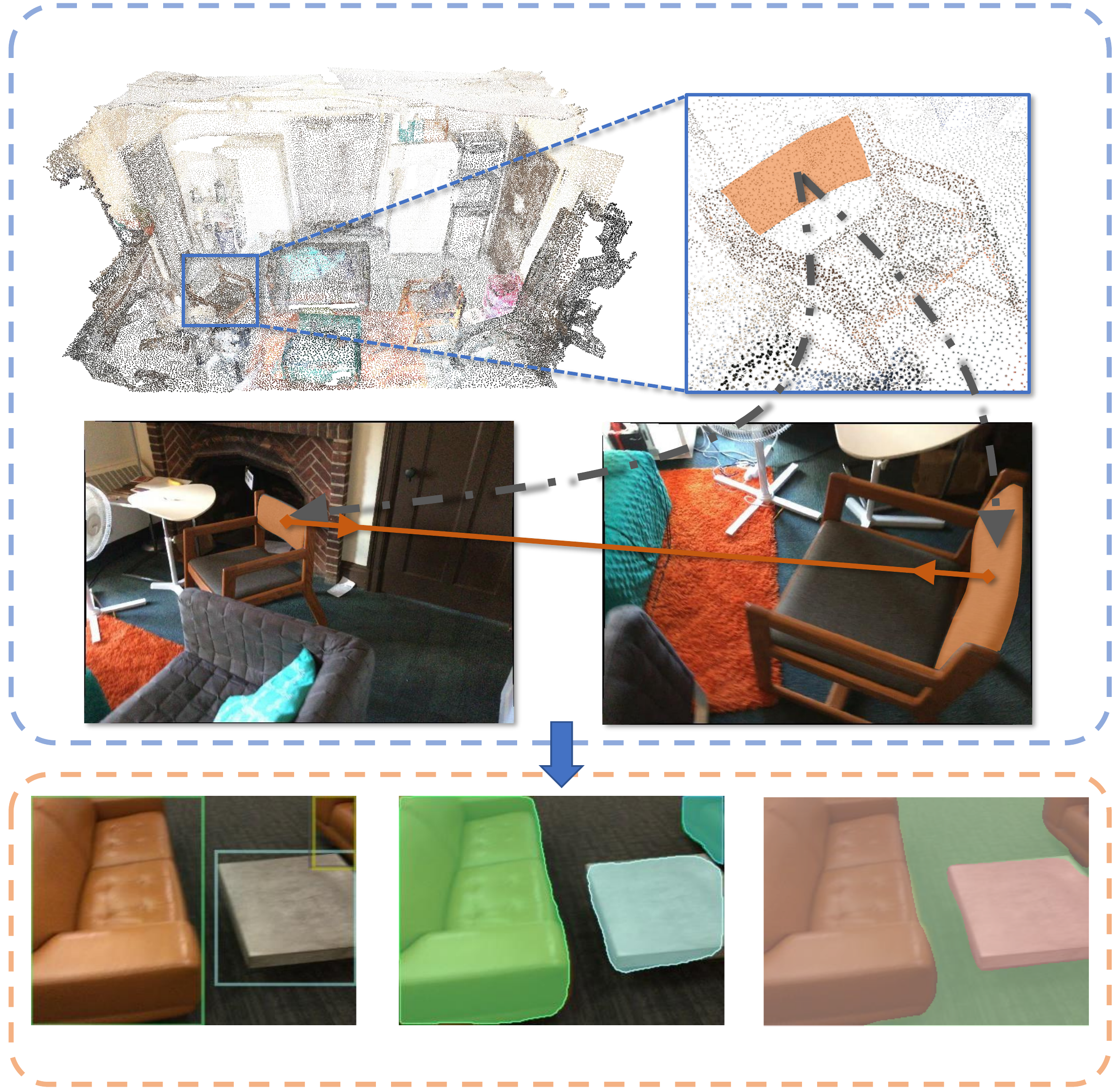}{
		\put(18,92.5){\small \textbf{Pre-training with Geometric Set Consistency}} 
		\put(6,3){\footnotesize Object Detection}
		\put(34,3){\footnotesize Instance Segmentation}
		\put(66,2.5){\footnotesize Semantic Segmentation}
    }\end{overpic}
    \vspace{-5mm}
  \caption{We introduce geometric consistency set to enforce the consistency within images for self-supervised image representation learning. We show that, the image representations pre-trained in this way can improve the performance of some 2D downstream tasks including semantic segmentation, object detection and instance segmentation.}
  \label{fig:teaser}
    \vspace{-5mm}
\end{figure}

Self-supervised image representation learning is an important problem in the field of computer vision and has been rapidly developed in recent years. Existing works in this area mainly focus on designing various pretext tasks to learn general and intrinsic image features in self-supervised manners \cite{oord2018representation,he2020momentum,hou2021pri3d}. Those pretext tasks are usually low-level and can capture general image properties that favor many downstream tasks, like image classification, semantic segmentation, object detection, instance segmentation, etc. Due to its ability to learn from a large amount of unlabeled data, self-supervised representation learning has already become a standard training regime in many real-world applications\cite{zhao2017pspnet,he2017mask,chen2017rethinking,chen2018encoder}.

Recently, researchers have started to use 3D data, usually represented by point clouds, meshes, or voxel grids, as guidance for learning image representations \cite{hou2021pri3d,liu2021contrastive}. Compared with 2D images, 3D data has complementary advantages for learning discriminative image features. Since 3D data is usually acquired by real-world scanning and reconstruction, and has the same dimension as the real-world scenes, learning geometric structures from 3D data is much easier than 2D images. Moreover, complex occlusions, as well as the texture of objects will also affect the performance of image perception methods.
Meanwhile, 3D data is occlusion-free, and geometric cues like smooth regions and sharp edges can be strong priors for semantic understanding. Thus it is natural to use 3D geometric cues to favor the learning of image representations.

Pioneering works, like Pri3D\cite{hou2021pri3d}, mainly rely on multi-view pixel-level consistency or 2D-3D pixel to point consistency for learning image representations in a self-supervised manner. 
The image representations learned in this way are proven to have significantly better performance than learning purely from 2D images for downstream tasks. Despite these great successes, the geometric consistency priors in 3D data (i.e. smooth regions or depth gaps) are not directly employed, which we demonstrate are strong cues and can significantly enhance the learning of image semantics.

In this paper, we propose to use geometric consistency to promote the learning of image representations. Our intuition is that 3D points within the same smooth or even planar regions may share similar semantics, while the discontinuities or depth gaps in 3D space may imply semantic changes. Such geometric cues are directly observable in unlabeled 3D data. However, due to complex textures, these cues can be hardly learned from purely unlabeled 2D images. Based on the above intuition, we design a simple yet effective method to learn the geometric consistency priors described above. Specifically, we leverage the continuities and discontinuities of 3D data, and use the clustering method to cluster the 3D data into many local small segments, termed \textit{geometric consistency sets}, to guide the learning of image representations in a self-supervised contrastive way. Our method is simple and can be easily implemented. We show that by exploring geometric consistency in the self-supervised pre-training stage, the performance of downstream tasks can be significantly improved.      


In the following, we summarize our main contributions:

1) We introduce geometric consistency into a contrastive learning framework for self-supervised image representation learning.

2) We propose a simple yet effective multi-view contrastive loss with geometric consistency sets to leverage the consistency within images.

3) We demonstrate superior performance on several downstream tasks compared with SOTA methods.

\section{Related Works}

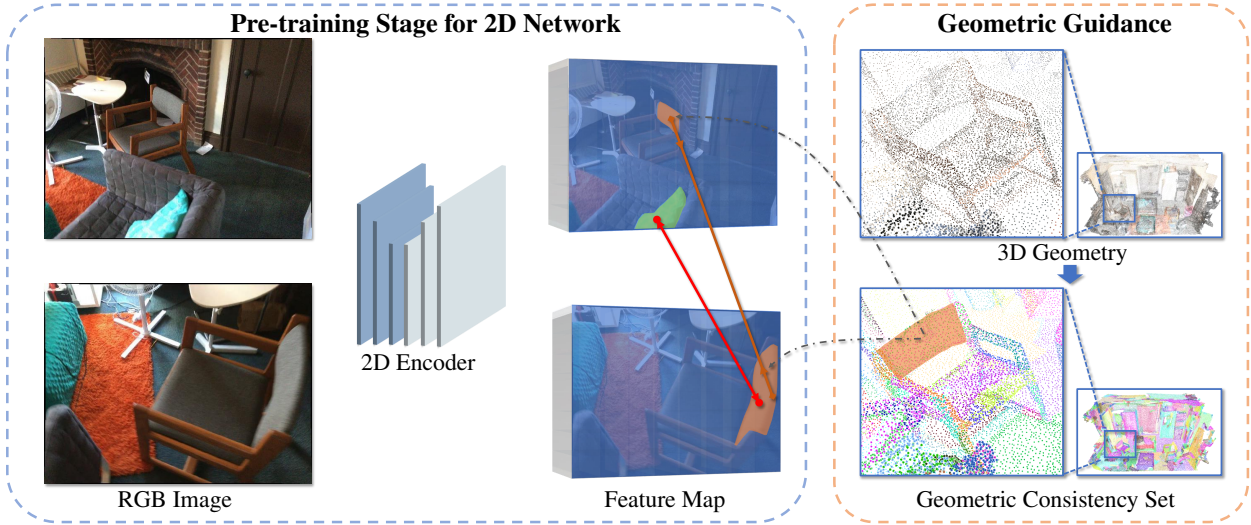
\begin{figure*}[ht]
\centering
\vspace{-5mm}
\begin{overpic}[width=0.95\linewidth]{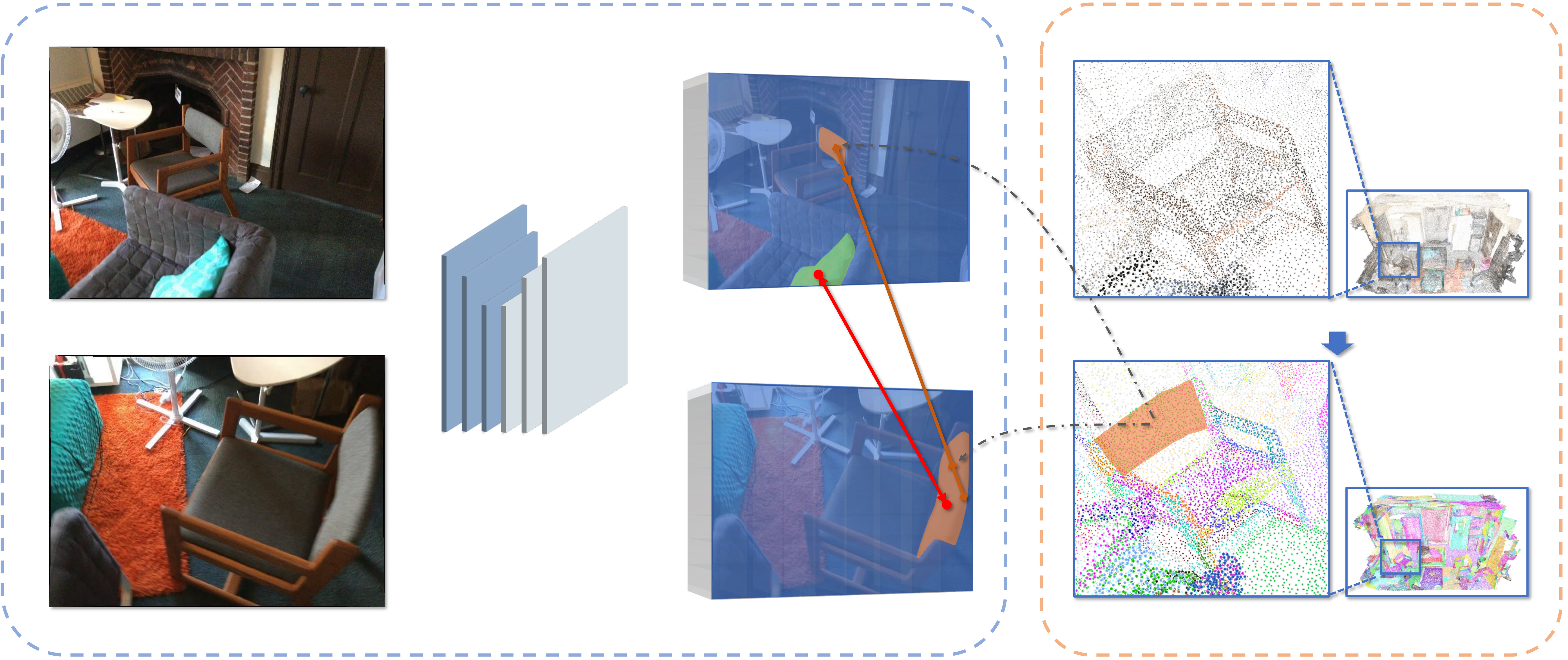}{
		\put(9,1.5){\small RGB Image} 
		\put(28.5,12.5){\small 2D Encoder} 
		\put(48,1.5){\small Feature Map} 
		\put(73,1.5){\small Geometric Consistency Set}
		\put(79.5,21.3){\small 3D Geometry}
		
		\put(17.5,39.6){ \textbf{Pre-training Stage for 2D Network}}
		\put(74.7,39.6){\textbf{Geometric Guidance}}
    }\end{overpic}
\vspace{-2mm}
  \caption{\textbf{Pipeline of our pre-training method}. Given a collection of multi-view unlabeled images together with the corresponding 3D geometry, we propose to use geometric consistency sets (illustrated with different colors), acquired via 3D clustering, to guide the learning of image representations. Specifically, the multi-view feature maps extracted with a 2D encoder will be first aggregated according to the corresponding geometric consistency sets, and then set-level InfoNCE loss will be used on the aggregated features to leverage the consistency within images. }
  \label{fig:pipeline}
\vspace{-5mm}
\end{figure*}

\subsection{Scene Understanding}
Learning based scene understanding tasks, including semantic scene segmentation \cite{qi2017pointnet,lin2020fpconv,liu20213d,hu2021bidirectional}, object detection \cite{hu2020jsenet,qi2020imvotenet,qi2019deep} and instance segmentation \cite{yi2019gspn,hou20193d,lahoud20193d}, are fundamental tasks in computer vision and have been rapidly developed in recent years. Due to the available of large scale, real scanned 2D and 3D datasets \cite{dai2017scannet,geiger2013vision,silberman2012indoor,armeni20163d}, these tasks have been widely explored. 

View-based methods \cite{dai20183dmv, kundu2020virtual} mainly rely on 2D images for scene understanding tasks, and the results produced with 2D networks can be further fused to 3D space based on multi-view consistency. In recent years, with the development of 3D deep learning networks \cite{qi2017pointnet, wu2019pointconv, wang2017cnn,choy20194d, graham20183d}, the performance of scene understanding tasks has been further promoted. The architectures of 3D networks can be roughly classified into two categories including point-based \cite{qi2017pointnet,wu2019pointconv}, and sparse-voxel based \cite{wang2017cnn,choy20194d,graham20183d}. These architectures are mainly designed for extracting features from sparse and unordered 3D data and have achieved great success in 3D scene understanding. Besides, since 2D images have clearer textures while 3D data is occlusion free from which extracting structure information is much easier, it is also an important research topic to study the joint training of 2D and 3D data \cite{song2018learning,liu2020p4contrast,hou2021pri3d,liu2021contrastive} for scene understanding tasks. 

In this paper, we propose to use 3D geometric consistency as guidance to promote the learning of 2D image representations and improve the performance of 2D scene understanding tasks.

\subsection{Self-Supervised Image Pre-training}
Self-supervised image pre-training as a fundamental way for learning representations from a large amount of data has shown significant impact in the field of computer vision and has been proved to be useful for various applications. Researchers in this field mainly aim at designing pretext tasks \cite{oord2018representation,larsson2017colorization,misra2020self,noroozi2016unsupervised} to learn intrinsic and general image representations that may favor various downstream tasks. Contrastive learning \cite{oord2018representation,he2020momentum,chen2020simple,chen2021exploring}, due to it's effectiveness and simplexity, has attracted great attention in recent years. The key idea of contrastive learning is to enforce the consistency between positive pairs while pushing away negative samples. Mainstream works mainly use various data augmentation strategies to get the positive pairs, including cropping, rotating, Gaussian blurring, etc. Besides, region-wise contrastive learning methods \cite{xiao2021region,liu2021bootstrapping} performance contrasting at region level to learn the region level similarities. Recently, researchers start to involve 3D data into contrastive pre-training \cite{hou2021pri3d,liu2020p4contrast,liu2021contrastive}, and the positive samples can be directly induced via multi-view consistency.   

In our work, we also explore using 3D data to promote the learning of image representations. In addition to multi-view consistency, we further leverage geometric consistency to enhance the consistency within image views.

\subsection{Multimodal Representation Learning}
Multimodal representation learning aims at learning joint representations by interacting between the data from different modalities. By incorporating the advantages from different modalities, the representations learned in this way are usually much better than those learned with a single modality only. Vision language pre-training \cite{radford2021learning,lu2019vilbert,li2020oscar} is a successful example. The availability of a large amount of image and language pairs makes the learned representations generalizable to various downstream tasks. Recently, with the availability of large scale RGB-D datasets, learning representations jointly from 2D and 3D data has attracted great attention in the research community \cite{song2018learning,liu2020p4contrast,hou2021pri3d,liu2021contrastive}. In particular, Pri3D \cite{hou2021pri3d} proposed to use 3D data as guidance for learning 2D to 2D and 2D to 3D pixel-level consistency. P4Contrast \cite{liu2020p4contrast} proposes to use point-pixel pairs for contrastive learning, and constructs negative samples artificially based on disturbed RGB-D points. TupleInfoNCE \cite{liu2021contrastive} compose new negative tuples for contrastive learning using modalities from different scenes to ensure that the weak modalities are not being ignored.

Existing works mainly use pixel-level multi-view consistency of single or multiply modalities. In our work, we use geometric consistency sets as guidance to explore intra-view consistency for 2D representation learning.


\subsection{Pseudo Labeling}
Pseudo labeling \cite{lee2013pseudo,shi2018transductive,rizve2021defense,liu2021one,iscen2019label}, as an effective way for learning with unlabeled data, has been widely studied in various applications. The key to the success of pseudo labeling methods is to generate high-quality pseudo labels for the unlabeled data. Researchers in this field have explored various ways for acquiring pseudo labels, including predicting directly with trained networks \cite{lee2013pseudo,liu2021one}, neighborhood graph propagation \cite{iscen2019label} or confidence based selection \cite{shi2018transductive,rizve2021defense} etc. 

The mainstream works mainly use pseudo labels in semi-supervised or weakly supervised learning scenarios. In our work, the geometric consistency sets we used serve as pseudo consistency labels to guide the learning of 2D image representations in a self-supervised manner.   
\section{Method}

Given a large collection of unlabeled images $\{I_{i}\}$ together with the corresponding 3D scenes $\{S_{i}\}$ represented as 3D points, the goal of our method is to pre-train an image encoder $f_{\theta}$ that can extract general and representative features from the input images with the guidance of geometric consistency in a self-supervised manner. And the pre-trained encoder $f_{\theta}$ can improve the performance of various downstream tasks after fine-tuning on relatively small labeled datasets. Note that 3D data is only used for pre-training and is not available in the fine-tuning stage.  

Figure \ref{fig:pipeline} illustrates our overall framework. In order to learn the image representations given a set of multi-view unlabeled images together with the corresponding 3D data, our observation is that the consistency or inconsistency of the geometric features in 3D space can serve as strong prior that may imply consistent semantics or object boundaries. For example, points that lie in the same smooth or even planar region may share the same semantics, while those separated by geometry dis-continuities or depth gaps may infer semantic changes. Based on the above observation, we propose to incorporate \textit{geometric consistency sets}, for example clustering the 3D data with similar geometry features into local smooth segments, into a contrastive learning framework to leverage the geometric consistency in learning image representations, and adapt the InfoNCE loss accordingly.  In the following, we will first describe the formulation of the \textit{set-InfoNCE loss} that generalizes the InfoNCE loss to enforce the consistency within image views. And then, we will provide the detailed implementation 
of our proposed contrastive training framework under the guidance of geometric consistency set. 

\subsection{Set-InfoNCE Loss}

\begin{figure}[ht]
\centering
\begin{overpic}[width=1.0\linewidth]{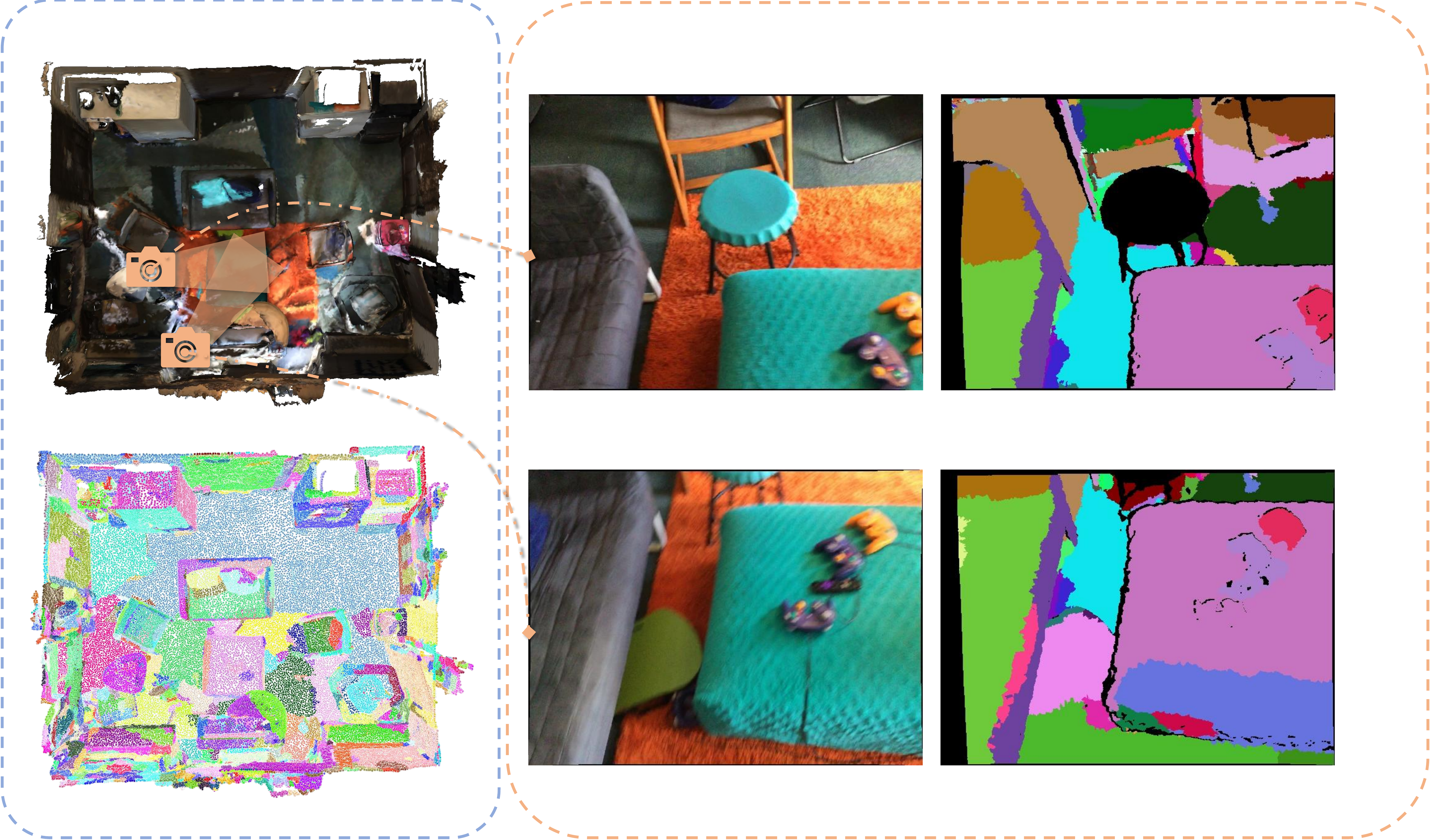}{

    \put(3,55){\footnotesize \bf{Geometric Guidance}}
    \put(58,55){\footnotesize \bf{2D Projection}}
    
    \put(8, 28){\footnotesize 3D Geometry}
    \put(1, 1.5){\footnotesize Geometric Consistency Set}
    
    \put(46,1.5){\footnotesize Image $I_n$}
    \put(46,28){\footnotesize Image $I_m$}
    
    \put(62,1.5){\footnotesize Projected Geo-sets $\{P_i^n\}_i$}
    \put(62,28){\footnotesize Projected Geo-sets $\{P_i^m\}_i$}

}\end{overpic}
    \vspace{-5mm}
  \caption{\textbf{Geometric consistency set.} We visualize the geometric consistency sets (with different colors) as well as the corresponding 2D projections on two corresponding image views. }
  \label{fig:pseudo_label}
\end{figure}
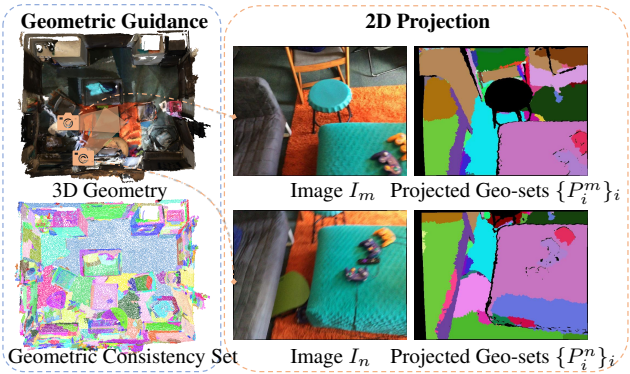

2D contrastive image representation learning mainly aims at learning consistent representations by contrasting multi-view images on either image level or pixel level. The multi-view images can be acquired via different data augmentation strategies or from real-world scanned sequences. Contrastive learning then enforces the feature consistency between different views that correspond to the same content while pushing away the others. For the pixel level contrastive representation learning, current works use pixel-level InfoNCE loss\cite{hou2021pri3d}, which is defined as:

\begin{equation}\label{eq:pcontrast_loss}
\mathcal{L}_{pixel} = - \sum_{(i,j) \in M_p } log \frac{\exp(\mathbf{f}_i \cdot \mathbf{f}_j / \tau)}{\sum_{(\cdot, k) \in M_p}{\exp(\mathbf{f}_i \cdot \mathbf{f}_k / \tau)}}
\end{equation}

\noindent where $M_p$ denotes the set of pixel-to-pixel matching pairs from one view to another, $\mathbf{f}_i$ is the normalized feature vector of the pixel at the $i$th position, and $\tau$ is the temperature parameter for controlling the concentration of the features in the representation space. 



Though the representations learned with pixel-level InfoNCE loss\cite{hou2021pri3d} has been proven to be beneficial for many downstream tasks, additional aggregated correspondences among the pixels like the geometric consistency prior could further improve the representation learning. In the following, we give the formulation of the set level InfoNCE loss that generalizes the pixel level InfoNCE to leverage such aggregated correspondences:


\begin{equation}\label{eq:set_contrast_loss}
\mathcal{L}_{set} = - \sum_{(i,j) \in M_s} log \frac{\exp(\mathcal{F}(P_i) \cdot \mathcal{F}(P_j) / \tau)}{\sum_{(\cdot,k)\in M_s}{\exp(\mathcal{F}(P_i) \cdot \mathcal{F}(P_k) / \tau)}} ,
\end{equation}

\noindent where $P_i$ is a set of feature points that are likely to have the same semantics, $\mathcal{F}$ is a mapping from a set to a feature vector in $\mathbb{R}^c$, and $M_s$ denotes the correspondence pairs among sets. Obviously Eqn.~\ref{eq:pcontrast_loss} is a special case to Eqn.~\ref{eq:set_contrast_loss} when the sets $P_i=\{\mathbf{f_i}\}$ degenerate to one-element only. 
In our implementation, we set $\mathcal{F}(P_i) = \frac{1}{|P_i|}\sum_{s_j\in P_i} \mathbf{f_{s_j}}$ for aggregating the features from all the points within sets, and we will also discuss different variations of $\mathcal{F}$ in Section \ref{diff_set_choice}.

Note that the set level InfoNCE loss defined in Eqn. \ref{eq:set_contrast_loss} is general, and the strategies for acquiring the set $P_i$ might be different for different scenarios \cite{xiao2021region,liu2021bootstrapping,wang2020unsupervised}. In this paper, we define our geometric consistency set under the guidance of 3D geometry, and will give the detailed description in Section \ref{sec:method_geosetconsistency}.

\subsection{Learning with Geometric Set Consistency} \label{sec:method_geosetconsistency}
Intuitively, the geometric consistency can be a strong prior that can guide the learning of within image consistency. 
The geometric cues like smooth regions and depth gaps may imply the same semantics or object boundaries respectively. In the following, we give the formal definition of the proposed set-InfoNCE loss with geometric consistency sets under projection.

Formally, for a given equivalence relation $\sim$ among the points in all the scenes $\cup S_i$, we call the quotient set $\{P_j\}=\cup S_i/\sim$ as the collection of \textit{geometric consistency sets}. For example, when we have some pre-defined geometric labels among the 3D points and define the equivalence relation as  \textit{points that share the same label}, then the geometric consistency set $P_i$ would contain all the points that have the $i$th label.  Besides, for each image $I_m$, we call $P_j^{m}=\{proj(s) \in I_m | s \in P_j\}$ as the $projection$ of $P_j$ from 3D onto 2D image view $I_m$. By adapting Eqn. \ref{eq:set_contrast_loss}, we could define our set-InfoNCE loss with geometric consistency set as:



\begin{equation}\label{eq:pseudo_loss}
\mathcal{L}_{geo\_set} = - \sum_{\mathclap{~(i,m,n) \in M_s~}} log \frac{\exp(\mathcal{F}(P^m_i) \cdot \mathcal{F}(P^n_i) / \tau)}{\sum_{(k, l, \cdot)\in M_s}{\exp(\mathcal{F}(P^m_i) \cdot \mathcal{F}(P^l_k) / \tau)}} ,
\end{equation}

\noindent  where $i$ is the index of the geometric consistency set and $m,n$ are the view indices. $M_s = \{(i,m,n)\}$ maintains the matching pairs of projections $P_i^m, P_i^n$ from one view to another for each geometric consistency set $P_i$.

In general, any proper notion of spatial equivalence induces its corresponding geometric consistency sets; in this work, we use a simple 3D clustering method and find it already works well in our case. 
We leave the exploration of more sophisticated spatial partitioning methods as future work.
Specifically, for a sequence of 2D images acquired by scanning around a specific scene, we first have the corresponding 3D surface $S$ via 3D reconstruction. To obtain geometric consistency sets, we use the surface over-segmentation results produced by a normal-based graph cut method \cite{felzenszwalb2004efficient,karpathy2013object}. 
As shown in Figure \ref{fig:pseudo_label}, the 3D points that are likely to have the same semantics can be clustered into the same geometric consistency set. 

\subsection{Training Strategy}

We use a two-stage training strategy to pre-train with our method. Besides our geometric guided set-InfoNCE loss that mainly considers set-level consistency and involves some high-level semantics such as smooth regions and surface discontinuities, we also include the pixel-level multi-view consistency to enforce the low-level distinction beneficial for the downstream tasks. Thus we propose to pre-train the network from low-level to high-level progressively with two stages. Specifically, in the first stage, we train the network with pixel to pixel and pixel to point multi-view contrastive loss termed as View and Geo loss in Pri3D\cite{hou2021pri3d}. And then, in the second stage, we continue to train the network with our geometric guided set-InfoNCE loss alone. In this way, we can achieve better performance on downstream tasks than training solely with our set-InfoNCE loss or with Pri3D losses.

\section{Experimental Setup}
In this section, we will provide a detailed description of our experimental setup, including the backbone architectures, datasets for pre-training and evaluation, and the implementation details. More experiment details are given in the supplementary materials.

\subsection{Backbone Architecture}

We use a UNet-style\cite{ronneberger2015u} architecture with residual connections as our network backbone since it has been widely used in various deep learning tasks. Specifically, it uses ResNet\cite{he2016deep} as encoder, and the decoder part contains standard convolution blocks together with bi-linear interpolation layers for upsampling. In most of our experiments, we use ResNet50 as our encoder for its efficiency and effectiveness. In addition, we also use ResNet18 as the encoder to see the effectiveness of our method with a more lightweight architecture.

\subsection{Dataset}

We conduct our experiments on two widely used datasets ScanNet\cite{dai2017scannet} and NYUv2 \cite{silberman2012indoor}, where ScanNet is used for both pre-training and downstream task fine-tuning, and NYUv2 is used for demonstrating the transferability of the representations learned with our method on ScanNet. In the following, we give more details about the two datasets.

\paragraph{ScanNet} 
ScanNet\cite{dai2017scannet} contains large amount of RGB-D sequences together with the corresponding 3D scene reconstructions. It has 1513 scenes for training, which contains around 2.5M images in total. Following Pri3D \cite{hou2021pri3d}, we regularly sample every 25th frame from the original ScanNet sequence for pre-training. And the frame pairs with more than 30$\%$ pixel overlap will be used for contrastive training. Totally, we have around 804k frame pairs in the pre-training stage. Note that we do not use any semantic labels for pre-training. For fine-tuning on downstream tasks, we use the standard labeled training and validation set of ScanNet benchmark\cite{dai2017scannet}, and the 2D images are sampled every 100th frame from the original sequence. Thus, in the fine-tuning stage, we have around 20k images for training and 5k images for validation.

\paragraph{NYUv2} NYUv2\cite{silberman2012indoor} is a dataset consisting of 2D images scanned from various indoor scenes, and in this paper, it is used for fine-tuning downstream tasks. Totally it has 1449 labeled images, in which 795 images are used for training and 654 images for testing.

\subsection{Comparison Methods}


\paragraph{Supervised ImageNet Pre-training \cite{deng2009imagenet}} The network is pre-trained with the supervised classification task on the ImageNet dataset.

\paragraph{MoCoV2 \cite{chen2020improved}}  For MoCoV2, there are various strategies for feeding the data, we only list the best performing result (i.e. MoCoV2-supIN $\rightarrow $ SN) from Pri3D \cite{hou2021pri3d}. To be specific, MoCoV2 is initialized with supervised ImageNet pre-training and then trained on ScanNet with randomly shuffled images.

\paragraph{Pri3D \cite{hou2021pri3d}}  Pri3D is the method most related to ours, in which 3D data and contrastive pre-training are used. Specifically, it comprises 2D to 2D and 2D to 3D point InfoNCE loss, termed View and Geo loss, respectively, and is also initialized with supervised ImageNet pre-training.

\paragraph{Depth Prediction}  The network is pre-trained with a single view depth prediction task, and is also initialized with supervised ImageNet pre-training. We obtain the results directly from Pri3D \cite{hou2021pri3d} where the method is implemented as a baseline.

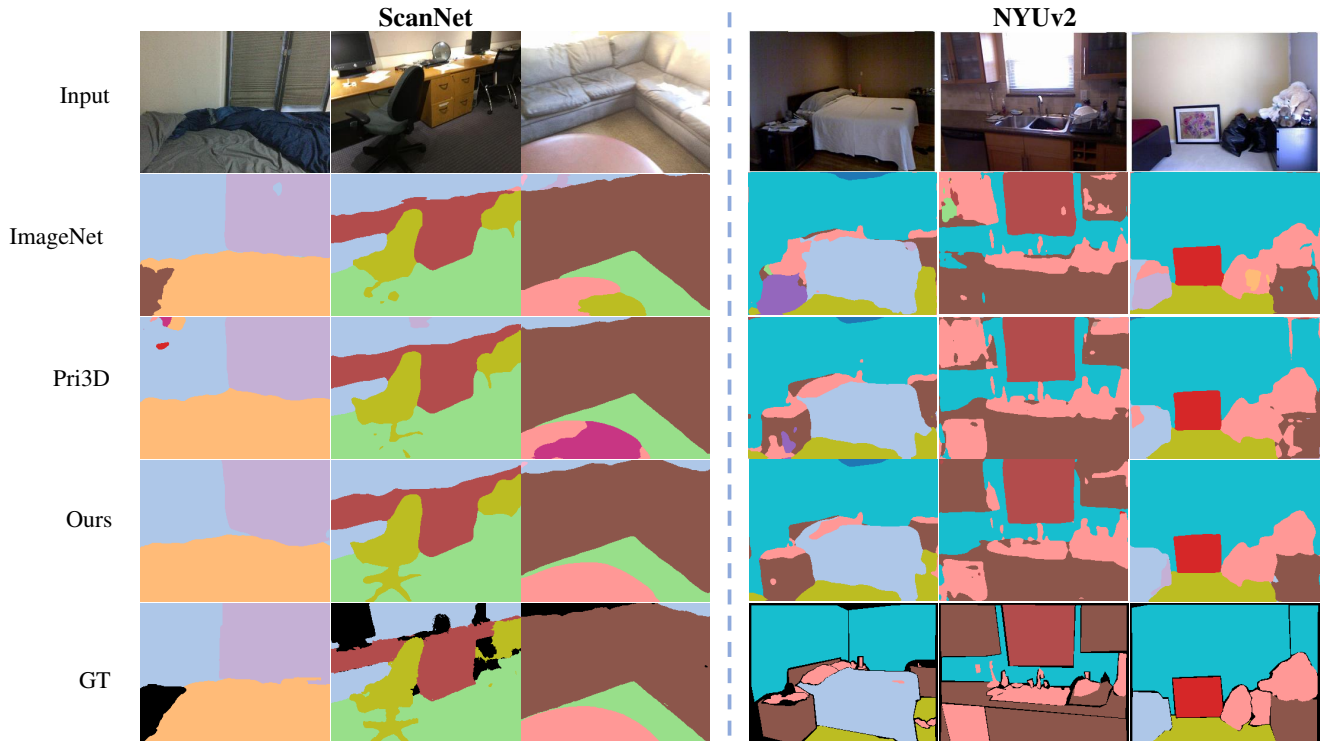
\begin{figure*}
\centering
\begin{overpic}[width=0.98\textwidth]{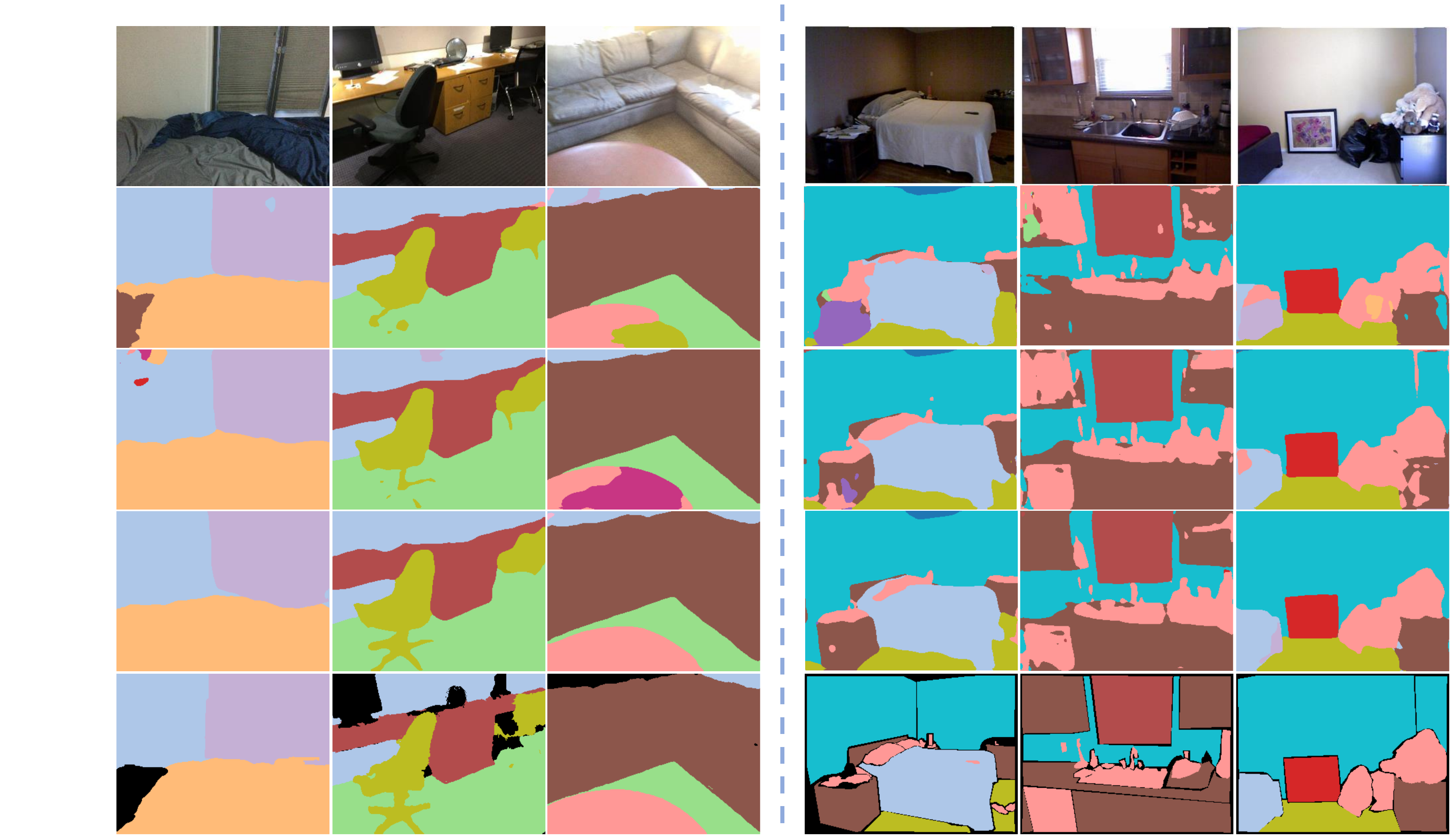}{
 		\put(26.5,56) {\textbf{ScanNet}}
 		\put(73.7,56){ \textbf{NYUv2}}
 		
 		\put(1.7,50){\small {Input}} 
 		\put(-2.2,39){\small {ImageNet}}
 		\put(1.2,28){\small {Pri3D}}
 		\put(2.2,17){\small {Ours}}
 		\put(3.2,4.5){\small {GT}}
    }\end{overpic}
\vspace{-3mm}
  \caption{\textbf{Qualitative results of semantic segmentation task on ScanNet and NYUv2 datasets.} All the methods are pre-trained on ScanNet with ResNet50 as the  backbone. For Pri3d we reproduce the results with their released code.}
  \label{fig:vis_semseg2d}
    
\end{figure*}

\subsection{Implementation Details}

We use the following setting in our experiments unless explicitly specified. For pre-training with our method, Stochastic gradient descent(SGD) \cite{robbins1951stochastic} with Polynomial Learning Rate Policy (PolyLR) \cite{shamir2013stochastic} is used for optimization. The initial learning rate is 0.1 and the batch size is 64. Following Pri3D\cite{hou2021pri3d}, we initialize the network with ImageNet pre-trained weights. In the first stage of pre-training, we train the network with View and Geo loss from Pri3D for 5 epochs. And then in the second stage, the network is trained with our proposed methods for 2 epochs. We use 8 NVIDIA V100 GPUs for training our network, and all the experiments are implemented with PyTorch.

\begin{table}
  \centering
  \begin{tabular}{lcc}
    \toprule
    Method & ResNet50 & ResNet18 \\
    \midrule
    From Scratch & 39.1 & 37.5 \\
    ImageNet Pre-training & 55.7 & 51.0 \\
    \midrule
    MoCoV2 & 56.6 & 52.9\\
    Depth Prediction & 58.4 & - \\
    Pri3D(View) & 61.3 & 54.4 \\
    Pri3D(Geo) & 61.1 & 55.3 \\
    Pri3D(View + Geo) & 61.7 & 55.7 \\
    \midrule
    Ours & \textbf{63.1} & \textbf{57.2} \\
    \bottomrule
  \end{tabular}
  \vspace{-2mm}
  \caption{\textbf{2D semantic segmentation results on ScanNet.} We use ResNet50 and ResNet18 as the backbone, mIoU for evaluation.}
  \label{tab:semseg2d_scannet}
\end{table}

\section{Experimental Results}

\subsection{Fine-tuning for 2D Downstream Tasks}
In the following, we demonstrate the effectiveness of our method by fine-tuning the pre-trained network on several 2D downstream tasks, including semantic segmentation, instance segmentation, and object detection.

\begin{table}
  \centering
  \begin{tabular}{lccccc}
    \toprule
    Method & 20\% & 40\% & 60\% & 80\% & 100\%\\
    \midrule
    Pri3d(View + Geo) & 51.5 & 56.2 & 58.9 & 60.3 & 61.7\\
    Ours & \textbf{54.1} & \textbf{58.3} & \textbf{60.4} & \textbf{61.5} & \textbf{63.1}\\
    \bottomrule
  \end{tabular}
  \vspace{-2mm}
  \caption{\textbf{Data efficient learning results on ScanNet.} We fine-tune our pre-trained representations with a limited amount of training data (from $20\%$ to $100\%$). ResNet50 is used as the backbone, and mIoU is used for evaluation. }
  \label{tab:semseg2d_scannet_limit}
\end{table}

\paragraph{Semantic Segmentation on ScanNet}

We demonstrate the effectiveness of our pre-trained representation by fine-tuning it with the ScanNet standard labeled training set, and mean IoU is used as the evaluation metric. ResNet50 or ResNet18 is used as the backbone encoder for all pre-training methods. In Table \ref{tab:semseg2d_scannet}, we compare our results with other state-of-the-art pre-training methods. Our method outperforms all the state-of-the-art methods for both ResNet50 and ResNet18 backbones, which confirms the effectiveness of our proposed self-supervised pre-training strategy with geometry-guided set consistency. Figure \ref{fig:vis_semseg2d} demonstrates some qualitative results, as we can see that the segmentation results of our method are less noisy and have clearer details.

We further test our method with different amounts of data for fine-tuning to study the performance under limited data scenarios. ResNet50 is used as the backbone. As shown in Table \ref{tab:semseg2d_scannet_limit}, by comparing our method with the best performing SOTA method, we can achieve superior performance consistently under different ratios of training data. 

Moreover, we also test our pre-trained representations with more advanced 2D semantic segmentation architectures including DeepLabV3 \cite{chen2017rethinking}, DeepLabV3+ \cite{chen2018encoder} and PSPNet\cite{zhao2017pspnet}. As shown in Table \ref{tab:semseg2d_scannet_diff_arch}, our method consistently outperforms the state-of-the-art methods by a large margin. Which further validate the effectiveness of considering geometric consistency set for self-supervised image representation learning.

\begin{table}
  \centering
  \begin{tabular}{lc}
    \toprule
    Method & ResNet50 \\
    \midrule
    DeepLabV3(ImageNet) & 57.0\\
    DeepLabV3(Pri3D) & 61.3\\
    DeepLabV3(Ours) & \textbf{62.2}\\
    \midrule
    DeepLabV3+(ImageNet) & 57.8\\
    DeepLabV3+(Pri3D) & 61.6\\
    DeepLabV3+(Ours) & \textbf{62.7}\\
    \midrule
    PSPNet(ImageNet) & 59.7\\
    PSPNet(Pri3D) & 62.8\\
    PSPNet(Ours) & \textbf{63.7}\\
    \bottomrule
  \end{tabular}
  \vspace{-2mm}
  \caption{\textbf{2D semantic segmentation results on ScanNet with different network architectures.} ResNet50 is used as the backbone for all architectures, and mIoU is used for evaluation.}
  \label{tab:semseg2d_scannet_diff_arch}
\end{table}

\paragraph{Instance Segmentation and Detection on ScanNet}

We fine-tune our pre-trained representations on 2D ScanNet object detection and instance segmentation tasks to see the generalizability of the learned representations. Specifically, ResNet50 is used as the backbone encoder for all the pre-training methods. Mask-RCNN \cite{he2017mask} implemented by Detectron2 \cite{wu2019detectron2} is used for object detection and instance segmentation tasks.  As shown in Table \ref{tab:insdet_scannet} and \ref{tab:insseg_scannet}, we can achieve comparable or superior performance compared with the other state-of-the-arts on different evaluation metrics. 

\begin{table}
  \centering
  \begin{tabular}{lccc}
    \toprule
    Method & AP@0.5 & AP@0.75 & AP \\
    \midrule
    From Scratch & 32.7 & 17.7 & 16.9 \\
    ImageNet Pre-training & 41.7 & 25.9 &  25.1\\
    \midrule
    MoCoV2 & 43.5 & 26.8 &  25.8\\
    Pri3D(View) & 43.7 & 27.0 &  26.3\\
    Pri3D(Geo) & 44.2 & \textbf{27.6} &  26.6\\
    Pri3D(View + Geo) & 44.5 & 27.4 &  26.6\\
    \midrule
    Ours & \textbf{45.1} & \textbf{27.6} &  \textbf{26.9}\\
    \bottomrule
  \end{tabular}
  \vspace{-2mm}
  \caption{\textbf{2D object detection results on ScanNet.} We use ResNet50 as the backbone and average precision for evaluation.}
  \label{tab:insdet_scannet}
\end{table}

\begin{table}
  \centering
  \begin{tabular}{lccc}
    \toprule
    Method & AP@0.5 & AP@0.75 & AP \\
    \midrule
    From Scratch & 25.8 & 13.1 & 12.2 \\
    ImageNet Pre-training & 32.6 & 17.8 &  17.6\\
    \midrule
    MoCoV2 & 33.9 & 18.1 &  18.3\\
    Pri3D(View) & 34.3 & 18.7 &  18.3\\
    Pri3D(Geo) & 34.4 & 18.7 &  18.3\\
    Pri3D(View + Geo) & 35.8 & \textbf{19.3} &  18.7\\
    \midrule
    Ours & \textbf{36.0} & \textbf{19.3} &  \textbf{19.5}\\

    \bottomrule
  \end{tabular}
  \vspace{-2mm}
  \caption{\textbf{2D instance segmentation results on ScanNet.} We use ResNet50 as the backbone and average precision for evaluation.}
  \label{tab:insseg_scannet}
\end{table}

\paragraph{Transfer to NYUv2}

We demonstrate the transferability of our pre-trained representations to other datasets. Specifically, we use ResNet50 as the backbone encoder for all pre-training methods.  The network is pre-trained on ScanNet\cite{dai2017scannet} dataset, and fine-tuned for downstream tasks on NYUv2\cite{silberman2012indoor} dataset. Following Pri3D\cite{hou2021pri3d}, we use the learning rate $0.01$ instead of $0.1$ for semantic segmentation task. As demonstrated in Table \ref{tab:semseg2d_nyuv2}, \ref{tab:insdet_nyuv2} and \ref{tab:insseg_nyuv2}, our pre-trained representations achieve superior performance compared with SOTA methods on most evaluation metrics, which further confirms that our pre-trained image representations are general and transferable across different datasets. 

\begin{table}
  \centering
  \begin{tabular}{lcc}
    \toprule
    Method & ResNet50 \\ 
    \midrule
    From Scratch & 24.8 \\
    ImageNet Pre-training & 50.0 \\
    \midrule
    MoCoV2 & 47.6 \\ 
    Pri3D(View) & 54.2 \\
    Pri3D(Geo) & 54.8 \\
    Pri3D(View + Geo) & 54.7 \\
    \midrule
    Ours & \textbf{55.4} \\
    \bottomrule
  \end{tabular}
  \vspace{-2mm}
  \caption{\textbf{2D semantic segmentation results on NYUv2.} mIoU is used as the evaluation metric.}
  \label{tab:semseg2d_nyuv2}
\end{table}

\begin{table}
  \centering
  \begin{tabular}{lccc}
    \toprule
    Method & AP@0.5 & AP@0.75 & AP \\
    \midrule
    From Scratch & 21.3 & 10.3 & 9.0 \\
    ImageNet Pre-training & 29.9 & 17.3 &  16.8\\
    \midrule
    MoCoV2 & 30.1 & 18.1 &  17.3\\
    Pri3D(View) & 33.0 & 19.8 &  18.9\\
    Pri3D(Geo) & 33.8 & 20.2 &  19.1\\
    Pri3D(View + Geo) & 34.0 & 20.4 &  19.4\\
    \midrule
    Ours & \textbf{34.6} & \textbf{20.5} &  \textbf{19.7}\\

    \bottomrule
  \end{tabular}
  \vspace{-2mm}
  \caption{\textbf{2D object detection results on NYUv2.} We use ResNet50 as the backbone, and average precision for evaluation.}
  \label{tab:insdet_nyuv2}
\end{table}

\begin{table}
  \centering
  \begin{tabular}{lccc}
    \toprule
    Method & AP@0.5 & AP@0.75 & AP \\
    \midrule
    From Scratch & 17.2 & 9.2 & 8.8 \\
    ImageNet Pre-training & 25.1 & 13.9 &  13.4\\
    \midrule
    MoCoV2 & 27.2 & 14.7 &  14.8\\
    Pri3D(View) & 28.1 & 15.7 &  15.7\\
    Pri3D(Geo) & 29.0 & 15.9 &  15.2\\
    Pri3D(View + Geo) & 29.5 &  \textbf{16.3} &  15.8\\
    \midrule
    Ours & \textbf{29.7} & \textbf{16.3} &  \textbf{16.5}\\

    \bottomrule
  \end{tabular}
  \vspace{-2mm}
  \caption{\textbf{2D instance segmentation results on NYUv2.} We use ResNet50 as the backbone, and average precision for evaluation.}
  \label{tab:insseg_nyuv2}
\end{table}

\subsection{Ablation Study}
\paragraph{Different Implementation of Set Feature Aggregation }\label{diff_set_choice}

We study a different choice of the set feature aggregation function $\mathcal{F}$ in Eqn. \ref{eq:pseudo_loss} to see whether it will influence the performance. Instead of using the average function to aggregate set-level features for both positive and negative pairs, here we test an asymmetric implementation. To be specific, for $\mathcal{F}(P^m_i)$ and $\mathcal{F}(P^l_k)$ in Eqn. \ref{eq:pseudo_loss} we select the feature of an arbitrary point from each set, and for $\mathcal{F}(P^n_i)$ average function is used. We pre-train the network with the newly defined loss on ScanNet and fine-tune it for ScanNet semantic segmentation. ResNet50 is used as the backbone. In this way, we get the performance of $63.1$ mIoU on the validation set, which is comparable with our result in Table \ref{tab:semseg2d_scannet}.

\paragraph{Ablation of Geometric Consistency Set}

We study the stability of our method by varying the parameters of the over-segmentation method \cite{felzenszwalb2004efficient,karpathy2013object} used for generating geometric consistency sets. To be specific, we generate geometric consistency sets on ScanNet under different clustering edge weight thresholds, and pre-train our network accordingly. As shown in Table ~\ref{tab:set_param_abl}, our method yields robustness to the clustering parameters.

\begin{table}[ht]
  \centering
  \scalebox{1}{
  \begin{tabular}{lccccc}
    \toprule
    Threshold & 0.01 & 0.02 & 0.03 & 0.04 & 0.05\\
    \midrule
    mIoU& 63.2 & 62.9 & 62.9 &63.2 & 63.1 \\
    \bottomrule
  \end{tabular}}
  \vspace{-1mm}
  \caption{\textbf{Ablation of geometric consistency set.} 2D semantic segmentation results on ScanNet with different over-segmentation parameters used for pre-train. ResNet50 is used as the backbone.}
  \label{tab:set_param_abl}
\end{table}

\paragraph{Pre-training with GT Semantic Label} Instead of using a geometry-guided consistency set, we also study the performance of the set-InfoNCE loss under the guidance of the ground truth semantic labels. Specifically, we use ground truth semantic labels to obtain the ground truth consistency sets and pre-train our network with ResNet50 as encoder on ScanNet\cite{dai2017scannet}. Note that ground truth category id is not used in pre-training. By fine-tuning the pre-trained representations for the ScanNet 2D semantic segmentation task, it could achieve the performance of 66.4 mIoU, which implies that a more sophisticated consistency set generation strategy may help to improve the result further. 

\paragraph{Ablation of Pri3D}

Since in the first stage of our method we use View and Geo loss in Pri3D\cite{hou2021pri3d} to train the initial representations, one may wonder whether training Pri3D for more epochs will lead to better performance compared with the results listed in their original paper. To address this concern, we test thoroughly by continuing training from the Pri3D checkpoint with different combinations of their losses for two more epochs. And then fine-tune the network on ScanNet semantic segmentation task\cite{dai2017scannet}.  Totally, in this experiment, Pri3D is pre-trained for 7 epochs. As shown in Table \ref{tab:pri3d_ablation}, continuous training of different configurations of Pri3D can not lead to better results.

\begin{table}
  \centering
  \begin{tabular}{lcc}
    \toprule
    Method & ResNet50\\
    \midrule
    Pri3D Original& 61.7 \\
    \midrule
    Pri3D Continue(View) & 61.4\\
    Pri3D Continue(Geo) & 61.6 \\
    Pri3D Continue(View + Geo) & 61.5  \\
    \midrule
    Ours & \textbf{63.1} \\
    \bottomrule
  \end{tabular}
  \vspace{-2mm}
  \caption{\textbf{Ablation of Pri3D.}  Continuous training of different configurations of Pri3D can not lead to better results. Pri3D Continue means continuous training of Pri3D.}
  \label{tab:pri3d_ablation}
\end{table}

\section{Limitations \& Future Work}
While our proposed pre-training method demonstrates superior performance compared with the state-of-the-art methods, there still exist several limitations. Since our approach relies on 3D geometric consistency to guide the learning of image representations, it can hardly be applied directly on the datasets such as ImageNet\cite{deng2009imagenet}, where 3D reconstructions are not applicable. Besides, in this work, we use a simple 3D clustering method for computing geometric consistency sets; one may consider using more advanced clustering techniques to further improve the performance.

\section{Conclusion}

We propose a method that uses 3D geometric consistency as guidance for self-supervised image representation learning. We leverage the continuities and discontinuities of 3D data and use the 3D clustering results to produce geometric consistency sets for
2D image views. Then we incorporate the geometric consistency sets into a contrastive learning framework to make the learned 2D image representations aware of 3D geometric consistency. We show that the image representations learned in this way will lead to superior performance than state-of-the-arts after fine-tuning for downstream tasks. Moreover, our approach can also improve the performance of downstream tasks when fine-tuning with limited training data. Various ablation studies further verify the effectiveness of our method.   


\normalem
{\small
\bibliographystyle{ieee_fullname}
\bibliography{ref}

\begin{thebibliography}{10}\itemsep=-1pt

\bibitem{armeni20163d}
Iro Armeni, Ozan Sener, Amir~R Zamir, Helen Jiang, Ioannis Brilakis, Martin
  Fischer, and Silvio Savarese.
\newblock 3d semantic parsing of large-scale indoor spaces.
\newblock In {\em Proceedings of the IEEE Conference on Computer Vision and
  Pattern Recognition}, pages 1534--1543, 2016.

\bibitem{chen2017rethinking}
Liang-Chieh Chen, George Papandreou, Florian Schroff, and Hartwig Adam.
\newblock Rethinking atrous convolution for semantic image segmentation.
\newblock {\em arXiv preprint arXiv:1706.05587}, 2017.

\bibitem{chen2018encoder}
Liang-Chieh Chen, Yukun Zhu, George Papandreou, Florian Schroff, and Hartwig
  Adam.
\newblock Encoder-decoder with atrous separable convolution for semantic image
  segmentation.
\newblock In {\em Proceedings of the European conference on computer vision
  (ECCV)}, pages 801--818, 2018.

\bibitem{chen2020simple}
Ting Chen, Simon Kornblith, Mohammad Norouzi, and Geoffrey Hinton.
\newblock A simple framework for contrastive learning of visual
  representations.
\newblock In {\em International conference on machine learning}, pages
  1597--1607. PMLR, 2020.

\bibitem{chen2020improved}
Xinlei Chen, Haoqi Fan, Ross Girshick, and Kaiming He.
\newblock Improved baselines with momentum contrastive learning.
\newblock {\em arXiv preprint arXiv:2003.04297}, 2020.

\bibitem{chen2021exploring}
Xinlei Chen and Kaiming He.
\newblock Exploring simple siamese representation learning.
\newblock In {\em Proceedings of the IEEE/CVF Conference on Computer Vision and
  Pattern Recognition}, pages 15750--15758, 2021.

\bibitem{choy20194d}
Christopher Choy, JunYoung Gwak, and Silvio Savarese.
\newblock 4d spatio-temporal convnets: Minkowski convolutional neural networks.
\newblock In {\em Proceedings of the IEEE/CVF Conference on Computer Vision and
  Pattern Recognition}, pages 3075--3084, 2019.

\bibitem{dai2017scannet}
Angela Dai, Angel~X. Chang, Manolis Savva, Maciej Halber, Thomas Funkhouser,
  and Matthias Nie{\ss}ner.
\newblock Scannet: Richly-annotated 3d reconstructions of indoor scenes.
\newblock In {\em Proc. Computer Vision and Pattern Recognition (CVPR), IEEE},
  2017.

\bibitem{dai20183dmv}
Angela Dai and Matthias Nie{\ss}ner.
\newblock 3dmv: Joint 3d-multi-view prediction for 3d semantic scene
  segmentation.
\newblock In {\em Proceedings of the European Conference on Computer Vision
  (ECCV)}, pages 452--468, 2018.

\bibitem{deng2009imagenet}
Jia Deng, Wei Dong, Richard Socher, Li-Jia Li, Kai Li, and Li Fei-Fei.
\newblock Imagenet: A large-scale hierarchical image database.
\newblock In {\em 2009 IEEE conference on computer vision and pattern
  recognition}, pages 248--255. Ieee, 2009.

\bibitem{felzenszwalb2004efficient}
Pedro~F Felzenszwalb and Daniel~P Huttenlocher.
\newblock Efficient graph-based image segmentation.
\newblock {\em International journal of computer vision}, 59(2):167--181, 2004.

\bibitem{geiger2013vision}
Andreas Geiger, Philip Lenz, Christoph Stiller, and Raquel Urtasun.
\newblock Vision meets robotics: The kitti dataset.
\newblock {\em The International Journal of Robotics Research},
  32(11):1231--1237, 2013.

\bibitem{graham20183d}
Benjamin Graham, Martin Engelcke, and Laurens Van Der~Maaten.
\newblock 3d semantic segmentation with submanifold sparse convolutional
  networks.
\newblock In {\em Proceedings of the IEEE conference on computer vision and
  pattern recognition}, pages 9224--9232, 2018.

\bibitem{he2020momentum}
Kaiming He, Haoqi Fan, Yuxin Wu, Saining Xie, and Ross Girshick.
\newblock Momentum contrast for unsupervised visual representation learning.
\newblock In {\em Proceedings of the IEEE/CVF Conference on Computer Vision and
  Pattern Recognition}, pages 9729--9738, 2020.

\bibitem{he2017mask}
Kaiming He, Georgia Gkioxari, Piotr Doll{\'a}r, and Ross Girshick.
\newblock Mask r-cnn.
\newblock In {\em Proceedings of the IEEE international conference on computer
  vision}, pages 2961--2969, 2017.

\bibitem{he2016deep}
Kaiming He, Xiangyu Zhang, Shaoqing Ren, and Jian Sun.
\newblock Deep residual learning for image recognition.
\newblock In {\em Proceedings of the IEEE conference on computer vision and
  pattern recognition}, pages 770--778, 2016.

\bibitem{hou20193d}
Ji Hou, Angela Dai, and Matthias Nie{\ss}ner.
\newblock 3d-sis: 3d semantic instance segmentation of rgb-d scans.
\newblock In {\em Proceedings of the IEEE/CVF Conference on Computer Vision and
  Pattern Recognition}, pages 4421--4430, 2019.

\bibitem{hou2021pri3d}
Ji Hou, Saining Xie, Benjamin Graham, Angela Dai, and Matthias Nie{\ss}ner.
\newblock Pri3d: Can 3d priors help 2d representation learning?
\newblock In {\em Proceedings of the IEEE/CVF International Conference on
  Computer Vision}, pages 5693--5702, 2021.

\bibitem{hu2021bidirectional}
Wenbo Hu, Hengshuang Zhao, Li Jiang, Jiaya Jia, and Tien-Tsin Wong.
\newblock Bidirectional projection network for cross dimension scene
  understanding.
\newblock In {\em Proceedings of the IEEE/CVF Conference on Computer Vision and
  Pattern Recognition}, pages 14373--14382, 2021.

\bibitem{hu2020jsenet}
Zeyu Hu, Mingmin Zhen, Xuyang Bai, Hongbo Fu, and Chiew-lan Tai.
\newblock Jsenet: Joint semantic segmentation and edge detection network for 3d
  point clouds.
\newblock In {\em Computer Vision--ECCV 2020: 16th European Conference,
  Glasgow, UK, August 23--28, 2020, Proceedings, Part XX 16}, pages 222--239.
  Springer, 2020.

\bibitem{iscen2019label}
Ahmet Iscen, Giorgos Tolias, Yannis Avrithis, and Ondrej Chum.
\newblock Label propagation for deep semi-supervised learning.
\newblock In {\em Proceedings of the IEEE/CVF Conference on Computer Vision and
  Pattern Recognition}, pages 5070--5079, 2019.

\bibitem{karpathy2013object}
Andrej Karpathy, Stephen Miller, and Li Fei-Fei.
\newblock Object discovery in 3d scenes via shape analysis.
\newblock In {\em 2013 IEEE International Conference on Robotics and
  Automation}, pages 2088--2095. IEEE, 2013.

\bibitem{kundu2020virtual}
Abhijit Kundu, Xiaoqi Yin, Alireza Fathi, David Ross, Brian Brewington, Thomas
  Funkhouser, and Caroline Pantofaru.
\newblock Virtual multi-view fusion for 3d semantic segmentation.
\newblock In {\em European Conference on Computer Vision}, pages 518--535.
  Springer, 2020.

\bibitem{lahoud20193d}
Jean Lahoud, Bernard Ghanem, Marc Pollefeys, and Martin~R Oswald.
\newblock 3d instance segmentation via multi-task metric learning.
\newblock In {\em Proceedings of the IEEE/CVF International Conference on
  Computer Vision}, pages 9256--9266, 2019.

\bibitem{larsson2017colorization}
Gustav Larsson, Michael Maire, and Gregory Shakhnarovich.
\newblock Colorization as a proxy task for visual understanding.
\newblock In {\em Proceedings of the IEEE Conference on Computer Vision and
  Pattern Recognition}, pages 6874--6883, 2017.

\bibitem{lee2013pseudo}
Dong-Hyun Lee et~al.
\newblock Pseudo-label: The simple and efficient semi-supervised learning
  method for deep neural networks.
\newblock In {\em Workshop on challenges in representation learning, ICML},
  volume~3, page 896, 2013.

\bibitem{li2020oscar}
Xiujun Li, Xi Yin, Chunyuan Li, Pengchuan Zhang, Xiaowei Hu, Lei Zhang, Lijuan
  Wang, Houdong Hu, Li Dong, Furu Wei, et~al.
\newblock Oscar: Object-semantics aligned pre-training for vision-language
  tasks.
\newblock In {\em European Conference on Computer Vision}, pages 121--137.
  Springer, 2020.

\bibitem{lin2020fpconv}
Yiqun Lin, Zizheng Yan, Haibin Huang, Dong Du, Ligang Liu, Shuguang Cui, and
  Xiaoguang Han.
\newblock Fpconv: Learning local flattening for point convolution.
\newblock In {\em Proceedings of the IEEE/CVF Conference on Computer Vision and
  Pattern Recognition}, pages 4293--4302, 2020.

\bibitem{liu2021bootstrapping}
Shikun Liu, Shuaifeng Zhi, Edward Johns, and Andrew~J Davison.
\newblock Bootstrapping semantic segmentation with regional contrast.
\newblock In {\em International Conference on Learning Representations}, 2022.

\bibitem{liu2021contrastive}
Yunze Liu, Qingnan Fan, Shanghang Zhang, Hao Dong, Thomas Funkhouser, and Li
  Yi.
\newblock Contrastive multimodal fusion with tupleinfonce.
\newblock In {\em Proceedings of the IEEE/CVF International Conference on
  Computer Vision}, pages 754--763, 2021.

\bibitem{liu2020p4contrast}
Yunze Liu, Li Yi, Shanghang Zhang, Qingnan Fan, Thomas Funkhouser, and Hao
  Dong.
\newblock P4contrast: Contrastive learning with pairs of point-pixel pairs for
  rgb-d scene understanding.
\newblock {\em arXiv preprint arXiv:2012.13089}, 2020.

\bibitem{liu20213d}
Zhengzhe Liu, Xiaojuan Qi, and Chi-Wing Fu.
\newblock 3d-to-2d distillation for indoor scene parsing.
\newblock In {\em Proceedings of the IEEE/CVF Conference on Computer Vision and
  Pattern Recognition}, pages 4464--4474, 2021.

\bibitem{liu2021one}
Zhengzhe Liu, Xiaojuan Qi, and Chi-Wing Fu.
\newblock One thing one click: A self-training approach for weakly supervised
  3d semantic segmentation.
\newblock In {\em Proceedings of the IEEE/CVF Conference on Computer Vision and
  Pattern Recognition}, pages 1726--1736, 2021.

\bibitem{lu2019vilbert}
Jiasen Lu, Dhruv Batra, Devi Parikh, and Stefan Lee.
\newblock Vilbert: Pretraining task-agnostic visiolinguistic representations
  for vision-and-language tasks.
\newblock {\em arXiv preprint arXiv:1908.02265}, 2019.

\bibitem{misra2020self}
Ishan Misra and Laurens van~der Maaten.
\newblock Self-supervised learning of pretext-invariant representations.
\newblock In {\em Proceedings of the IEEE/CVF Conference on Computer Vision and
  Pattern Recognition}, pages 6707--6717, 2020.

\bibitem{noroozi2016unsupervised}
Mehdi Noroozi and Paolo Favaro.
\newblock Unsupervised learning of visual representations by solving jigsaw
  puzzles.
\newblock In {\em European conference on computer vision}, pages 69--84.
  Springer, 2016.

\bibitem{oord2018representation}
Aaron van~den Oord, Yazhe Li, and Oriol Vinyals.
\newblock Representation learning with contrastive predictive coding.
\newblock {\em arXiv preprint arXiv:1807.03748}, 2018.

\bibitem{papon2013voxel}
Jeremie Papon, Alexey Abramov, Markus Schoeler, and Florentin Worgotter.
\newblock Voxel cloud connectivity segmentation-supervoxels for point clouds.
\newblock In {\em Proceedings of the IEEE conference on computer vision and
  pattern recognition}, pages 2027--2034, 2013.

\bibitem{qi2020imvotenet}
Charles~R Qi, Xinlei Chen, Or Litany, and Leonidas~J Guibas.
\newblock Imvotenet: Boosting 3d object detection in point clouds with image
  votes.
\newblock In {\em Proceedings of the IEEE/CVF conference on computer vision and
  pattern recognition}, pages 4404--4413, 2020.

\bibitem{qi2019deep}
Charles~R Qi, Or Litany, Kaiming He, and Leonidas~J Guibas.
\newblock Deep hough voting for 3d object detection in point clouds.
\newblock In {\em Proceedings of the IEEE/CVF International Conference on
  Computer Vision}, pages 9277--9286, 2019.

\bibitem{qi2017pointnet}
Charles~R Qi, Hao Su, Kaichun Mo, and Leonidas~J Guibas.
\newblock Pointnet: Deep learning on point sets for 3d classification and
  segmentation.
\newblock In {\em Proceedings of the IEEE conference on computer vision and
  pattern recognition}, pages 652--660, 2017.

\bibitem{radford2021learning}
Alec Radford, Jong~Wook Kim, Chris Hallacy, Aditya Ramesh, Gabriel Goh,
  Sandhini Agarwal, Girish Sastry, Amanda Askell, Pamela Mishkin, Jack Clark,
  et~al.
\newblock Learning transferable visual models from natural language
  supervision.
\newblock In {\em International Conference on Machine Learning}, pages
  8748--8763. PMLR, 2021.

\bibitem{rizve2021defense}
Mamshad~Nayeem Rizve, Kevin Duarte, Yogesh~S Rawat, and Mubarak Shah.
\newblock In defense of pseudo-labeling: An uncertainty-aware pseudo-label
  selection framework for semi-supervised learning.
\newblock In {\em International Conference on Learning Representations}, 2021.

\bibitem{robbins1951stochastic}
Herbert Robbins and Sutton Monro.
\newblock A stochastic approximation method.
\newblock {\em The annals of mathematical statistics}, pages 400--407, 1951.

\bibitem{ronneberger2015u}
Olaf Ronneberger, Philipp Fischer, and Thomas Brox.
\newblock U-net: Convolutional networks for biomedical image segmentation.
\newblock In {\em International Conference on Medical image computing and
  computer-assisted intervention}, pages 234--241. Springer, 2015.

\bibitem{Rusu_ICRA2011_PCL}
Radu~Bogdan Rusu and Steve Cousins.
\newblock {3D is here: Point Cloud Library (PCL)}.
\newblock In {\em {IEEE International Conference on Robotics and Automation
  (ICRA)}}, Shanghai, China, May 9-13 2011.

\bibitem{shamir2013stochastic}
Ohad Shamir and Tong Zhang.
\newblock Stochastic gradient descent for non-smooth optimization: Convergence
  results and optimal averaging schemes.
\newblock In {\em International conference on machine learning}, pages 71--79.
  PMLR, 2013.

\bibitem{shi2018transductive}
Weiwei Shi, Yihong Gong, Chris Ding, Zhiheng~MaXiaoyu Tao, and Nanning Zheng.
\newblock Transductive semi-supervised deep learning using min-max features.
\newblock In {\em Proceedings of the European Conference on Computer Vision
  (ECCV)}, pages 299--315, 2018.

\bibitem{silberman2012indoor}
Nathan Silberman, Derek Hoiem, Pushmeet Kohli, and Rob Fergus.
\newblock Indoor segmentation and support inference from rgbd images.
\newblock In {\em European conference on computer vision}, pages 746--760.
  Springer, 2012.

\bibitem{song2015sun}
Shuran Song, Samuel~P Lichtenberg, and Jianxiong Xiao.
\newblock Sun rgb-d: A rgb-d scene understanding benchmark suite.
\newblock In {\em Proceedings of the IEEE conference on computer vision and
  pattern recognition}, pages 567--576, 2015.

\bibitem{song2018learning}
Xinhang Song, Shuqiang Jiang, Luis Herranz, and Chengpeng Chen.
\newblock Learning effective rgb-d representations for scene recognition.
\newblock {\em IEEE Transactions on Image Processing}, 28(2):980--993, 2018.

\bibitem{wang2017cnn}
Peng-Shuai Wang, Yang Liu, Yu-Xiao Guo, Chun-Yu Sun, and Xin Tong.
\newblock O-cnn: Octree-based convolutional neural networks for 3d shape
  analysis.
\newblock {\em ACM Transactions On Graphics (TOG)}, 36(4):1--11, 2017.

\bibitem{wang2020unsupervised}
Peng-Shuai Wang, Yu-Qi Yang, Qian-Fang Zou, Zhirong Wu, Yang Liu, and Xin Tong.
\newblock Unsupervised {3D} learning for shape analysis via multiresolution
  instance discrimination, 2021.

\bibitem{wu2019pointconv}
Wenxuan Wu, Zhongang Qi, and Li Fuxin.
\newblock Pointconv: Deep convolutional networks on 3d point clouds.
\newblock In {\em Proceedings of the IEEE/CVF Conference on Computer Vision and
  Pattern Recognition}, pages 9621--9630, 2019.

\bibitem{wu2019detectron2}
Yuxin Wu, Alexander Kirillov, Francisco Massa, Wan-Yen Lo, and Ross Girshick.
\newblock Detectron2.
\newblock \url{https://github.com/facebookresearch/detectron2}, 2019.

\bibitem{xiao2021region}
Tete Xiao, Colorado~J Reed, Xiaolong Wang, Kurt Keutzer, and Trevor Darrell.
\newblock Region similarity representation learning.
\newblock In {\em Proceedings of the IEEE/CVF International Conference on
  Computer Vision}, pages 10539--10548, 2021.

\bibitem{yi2019gspn}
Li Yi, Wang Zhao, He Wang, Minhyuk Sung, and Leonidas~J Guibas.
\newblock Gspn: Generative shape proposal network for 3d instance segmentation
  in point cloud.
\newblock In {\em Proceedings of the IEEE/CVF Conference on Computer Vision and
  Pattern Recognition}, pages 3947--3956, 2019.

\bibitem{yu2020learning}
Yaodong Yu, Kwan Ho~Ryan Chan, Chong You, Chaobing Song, and Yi Ma.
\newblock Learning diverse and discriminative representations via the principle
  of maximal coding rate reduction.
\newblock {\em Advances in Neural Information Processing Systems},
  33:9422--9434, 2020.

\bibitem{zhao2017pspnet}
Hengshuang Zhao, Jianping Shi, Xiaojuan Qi, Xiaogang Wang, and Jiaya Jia.
\newblock Pyramid scene parsing network.
\newblock In {\em CVPR}, 2017.

\end{thebibliography}
}

\clearpage
\newpage

\begin{appendices}

\section{Details of Geometric Consistency Set}

\begin{figure*}
\centering
\begin{overpic}[width=0.95\textwidth]{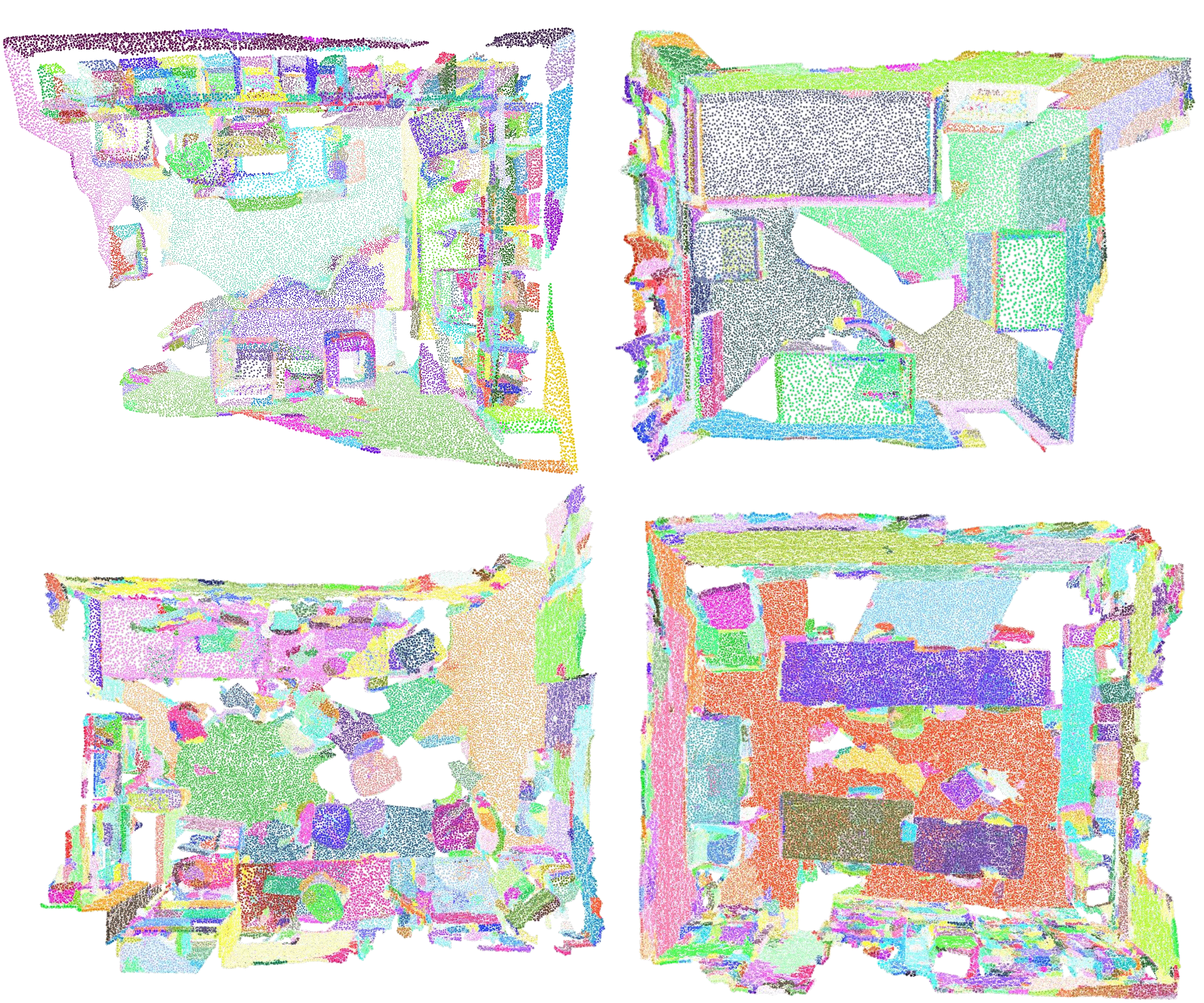}{
}\end{overpic}

  \caption{\textbf{Geometric consistency sets on 3D.} Different colors indicate different geometric consistency sets. }
  \label{fig:vis_patches}
\end{figure*}

\begin{figure*}
\centering
\begin{overpic}[width=1\textwidth]{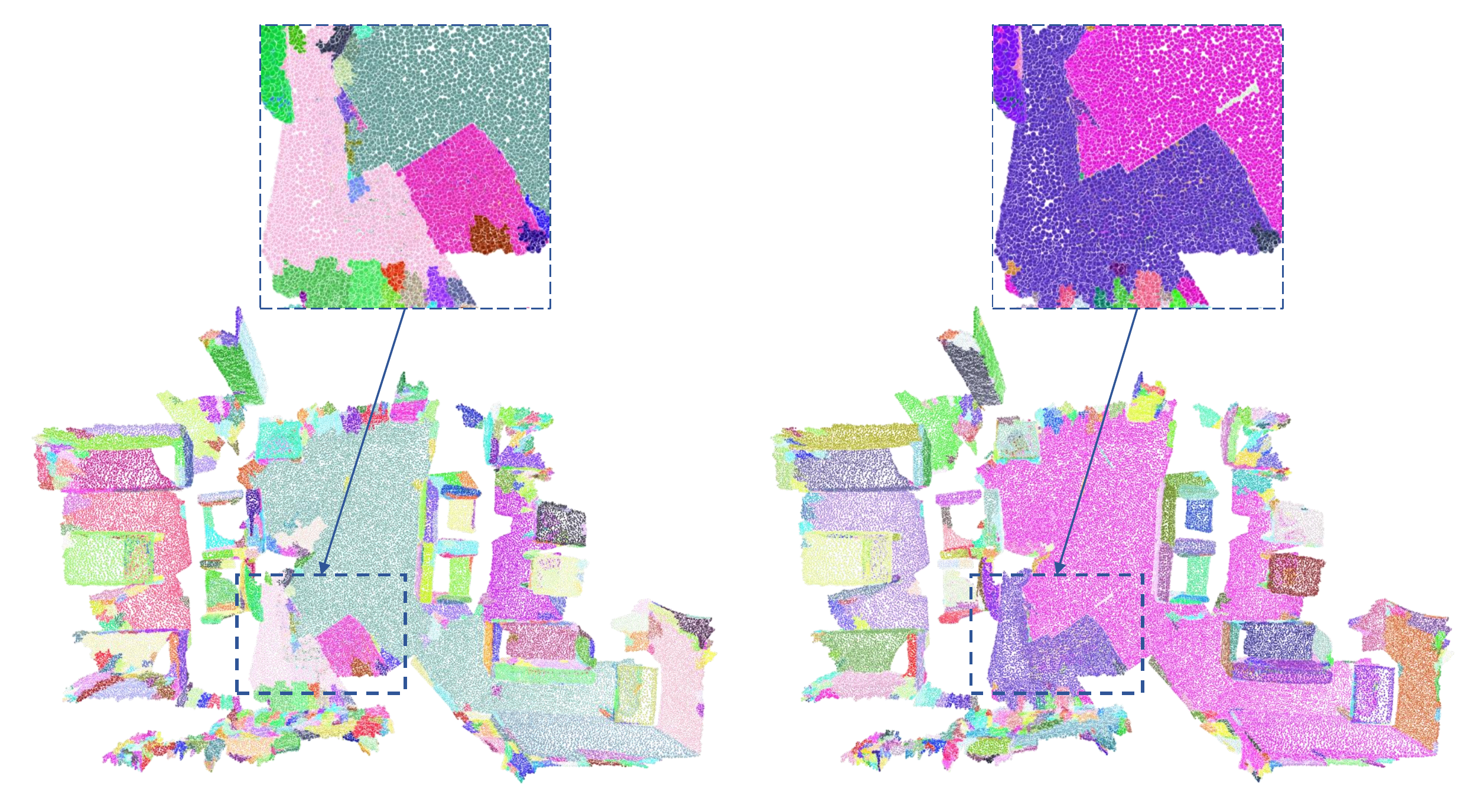}{

	\put(3,45){\footnotesize Threshold: 0.01}
	\put(52,45){\footnotesize Threshold: 0.05}
}\end{overpic}

  \caption{\textbf{3D geometric consistency sets generated by different parameters.}   Different colors indicate different geometric consistency sets, and larger clustering edge weight threshold will lead to larger sets. Black color denotes the unlabeled regions.}
  \label{fig:vis_geo_sets_param}

\end{figure*}

Here we provide more details about how to acquire the geometric consistency sets. For a 3D scene surface $S$ we can have a sequence of corresponding RGB-D scanning images. We use the surface over-segmentation results produced by a normal-based graph cut method ~\cite{felzenszwalb2004efficient, karpathy2013object} as the geometric consistency sets $\{P_j\}$. In this way, the 3D scene will be divided into many small segments. Figure \ref{fig:vis_patches} shows some examples, and Figure \ref{fig:vis_geo_sets_param} demonstrates the sets generated with different clustering edge weight thresholds. Then we project the 3D surface points together with the corresponding geometric consistency set id (here we use different ids to label the points in different geometric consistency sets) from 3D to 2D image views, and the pinhole camera model is used for 3D to 2D projection. Since the matching between the reconstructed 3D surface and the corresponding 2D image views may have a miss-match problem, we filter out invalid projections by comparing the depth difference between the projection points and the view depth maps. The projected points with the depth difference larger than a threshold, 0.05 in our experiments, will be regarded as invalid. Based on the valid projection points, we can have the corresponding projection $P_j^{m}=\{proj(s) \in I_m | s \in P_j\}$ of the geometric consistency set $P_j$ from 3D onto 2D image view $I_m$.

\section{Effect of the Initial Learning Rate}

We study different choices of the initial learning rate in the pre-training stage to see how it will influence the fine-tuning results. Specifically, the networks are pre-trained with different initial learning rates, including $0.1$ and $0.01$, on the ScanNet \cite{dai2017scannet} dataset and fine-tuned for the image semantic segmentation task on ScanNet and NYUv2 \cite{silberman2012indoor} datasets.  Table \ref{tab:learning_rate} illustrates the performance of both our method and Pri3D \cite{hou2021pri3d}. Although the network performance varies with the initial learning rate, our method consistently outperforms Pri3D on all the settings.

\begin{table}[hbt!]
  \centering
  \begin{tabular}{lcc}
    \toprule
    Method & ScanNet & NYUv2\\
    \midrule
    Pri3D (0.01) & 59.7 & 54.8  \\
    Ours (0.01) & \textbf{60.3} &  \textbf{55.4}  \\
    \midrule
    Pri3D (0.1) & 61.7 & 51.4 \\
    Ours (0.1) & \textbf{63.1} & \textbf{54.1}\\
    \bottomrule
  \end{tabular}
  \caption{\textbf{Effect of the initial learning rate.}  Effect of different initial learning rates in the pre-training stage. ResNet50 is used as the backbone encoder, the network is pre-trained on ScanNet dataset and fine-tuned for the image semantic segmentation task on ScanNet and NYUv2 datasets. mIOU is used for evaluation.}
  \label{tab:learning_rate}
  
\end{table}

 \section{Convergence}

\begin{figure}[ht]
\centering
\begin{overpic}[width=0.45\textwidth]{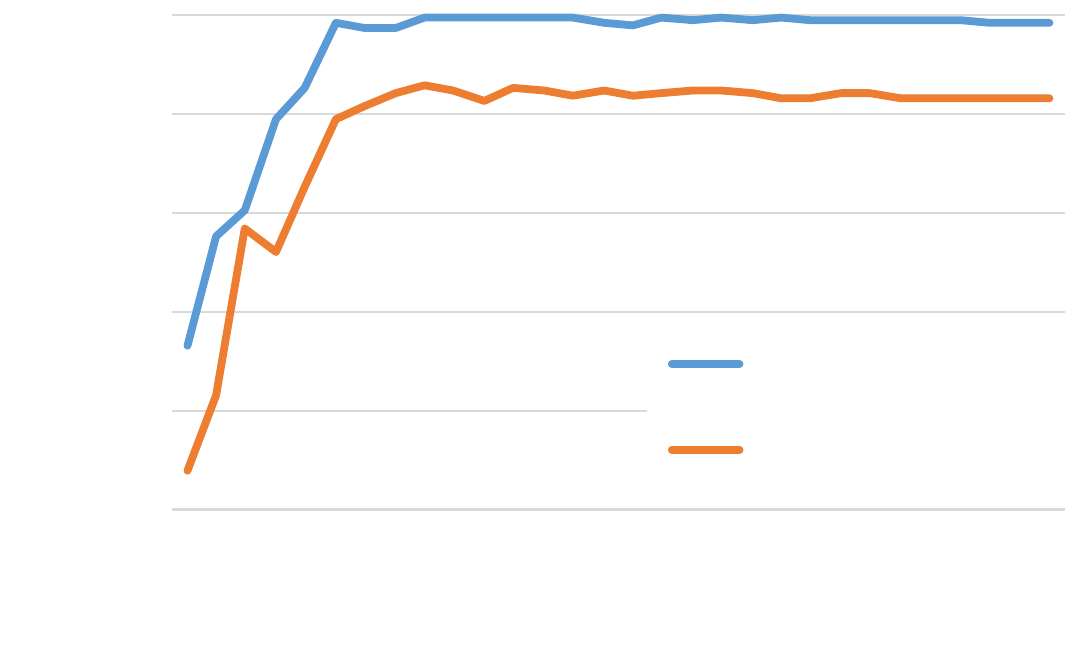}{
 		\put(16,6){\footnotesize 1}
 		\put(28,6){\footnotesize 5}
 		\put(40,6){\footnotesize 10}
 		\put(54,6){\footnotesize 15}
 		\put(67.5,6){\footnotesize 20}
 		\put(82,6){\footnotesize 25}
 		\put(96,6){\footnotesize 30}
 		
 		\put(36,1){\footnotesize Number of Fine-tuning Epochs}
 		
 		\put(70, 26){\footnotesize Ours}
 		\put(70, 18){\footnotesize Pri3d(View+Geo)}
 		
 		\put(6,12){\footnotesize 0.53}
 		\put(6,22){\footnotesize 0.55}
 		\put(6,31){\footnotesize 0.57}
 		\put(6,40){\footnotesize 0.59}
 		\put(6,50){\footnotesize 0.61}
 		\put(6,59){\footnotesize 0.63}
 		
 		\put(-5, 35){\footnotesize mIoU}
    }\end{overpic}
    
  \caption{\textbf{Convergence of the methods.} We fine-tune the pre-trained networks on the ScanNet semantic segmentation task. The average performance of 3 runs for each method is reported.}
  \label{fig:convergence}

\end{figure}

We study the convergence of the methods by fine-tuning the pre-trained networks on ScanNet\cite{dai2017scannet} semantic segmentation dataset and reporting the performance on the validation set after different number of epochs. As shown in Figure \ref{fig:convergence}, both Pri3D\cite{hou2021pri3d} and our proposed method converge within $10$ epochs, and our method consistently outperforms Pri3D after that.   

\section{Representation Space Analysis}

We analyze the quality of the learned representation spaces by computing the coding rate \cite{yu2020learning}, which measures the intra-category compactness, on ScanNet 2D semantic segmentation validation set.

\begin{figure*}[h]

\centering

\begin{overpic}[width=1\textwidth]{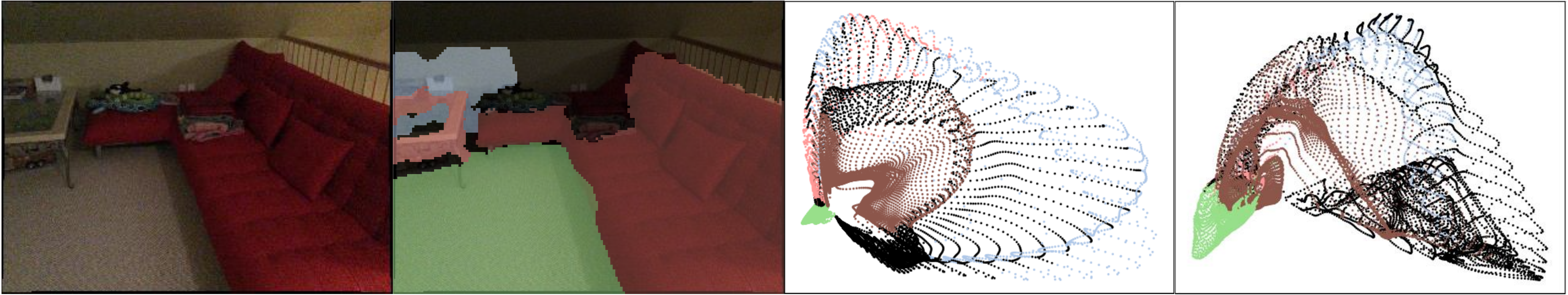}{

		\put(86,-2){\footnotesize Pri3D}
		\put(61,-2){\footnotesize Ours}
		\put(35,-2){\footnotesize GT Label}
		\put(10,-2){\footnotesize Input}

    }\end{overpic}

    \vspace{4mm}
  \caption{\textbf{PCA embedding of learned pixel features.} Different colors indicate different ground truth segmentation categories, and  black color indicates the unlabeled regions.}
  \label{fig:vis_feat}

\end{figure*}

Specifically, let  $\mathbf{F} \in \mathbb{R}^{d \times m}$ be the matrix containing $m$ feature vectors with dimension $d$. The coding rate of $\mathbf{F}$ can be defined as:

\begin{equation}\label{eq:coding_rate}
R(\mathbf{F}, \epsilon) = \frac{1}{2} \log \det (\mathbf{I} + \frac{d}{m \epsilon^2} \mathbf{F}\mathbf{F}^{\mathsf{T}})
\end{equation}

\noindent where $I$ is the identity matrix, and $\epsilon$ is the distortion parameter. 
For each image, we extract pixel features with pre-trained networks. Since the pixel features extracted with different pre-trained networks may have different overall scales, we scale the features by dividing by average feature length. Then, we compute the coding rate for the features within each ground truth category. The coding rate of an image can be computed by averaging the coding rates of all the categories within this image. The average coding rates of Pri3D \cite{hou2021pri3d} and ours are $54.04$ and $34.05$ respectively. This means our pre-trained representations are more compact than Pri3D. Moreover, we also visualize the learned features by PCA. As shown in Figure ~\ref{fig:vis_feat}, our features are cleaner and more separable.



\section{Ablation of Two-stage Training} 
We test our method with different pre-training configurations, i.e. set-InfoNCE loss only, set-InfoNCE plus pixel-InfoNCE  and two-stage training. As shown in Table ~\ref{tab:two_stage_abl}, the best performance is achieved with two-stage training that learns from low-level to high-level. Similar strategies can also be found in research topics like curriculum learning.

\begin{table}[ht]

  \centering
  \scalebox{0.97}{
  \begin{tabular}{lccccc}
    \toprule
     &  set-InfoNCE & set + pixel-InfoNCE & two stage \\
    \midrule
    mIoU& 60.6 & 61.0 & 63.1 \\
    \bottomrule
  \end{tabular}}

  \caption{\textbf{Two-stage training ablation. }Performance of 2D semantic segmentation task on ScanNet with different pre-training configurations. ResNet50 is used as the backbone encoder.}
  \label{tab:two_stage_abl}
\end{table}

\section{Performance on SUN RGB-D Dataset}
To further validate the transferability of our method, we fine-tune the ScanNet pre-trained representations on SUN RGB-D Dataset \cite{song2015sun} for the 2D semantic segmentation task. Specifically, the dataset contains 5k images for training and 5k images for testing, and the networks are pre-trained with initial learning rates of $0.1$ and $0.01$ respectively. As shown in Table ~\ref{tab:transfer_sunrgbd}, our method achieves better performance compared with Pri3D \cite{hou2021pri3d}.

\begin{table}[ht]
 
  \centering
  \scalebox{1}{
    \begin{tabular}{lcc}
    \toprule
    Method & ResNet50 \\
    \midrule
    ImageNet Pre-training & 34.8\\
    Pri3D (0.1) & 37.3   \\
    Ours (0.1) & 38.1  \\
    Pri3D (0.01) & 38.6\\
    Ours (0.01) & \textbf{39.2}\\
   
    \bottomrule
  \end{tabular}}

  \caption{\textbf{Performance on SUN RGB-D dataset.} Pre-train on ScanNet with different learning rates and fine-tune on SUN RGB-D for 2D semantic segmentation. ResNet50 is used as the backbone encoder.}
  \label{tab:transfer_sunrgbd}
\end{table}

\begin{table}[hbt!]
  \centering
  \begin{tabular}{lcc}
    \toprule
    Method & ResNet50 \\
    \midrule
    ImageNet Pre-training & 28.5   \\
    Pri3D & 33.2  \\
    \midrule
    Ours  & \textbf{33.7} \\
    \bottomrule
  \end{tabular}
  \caption{\textbf{Performance on KITTI dataset.}  ResNet50 is used as the backbone encoder, the network is pre-trained with unlabeled RGB-D sequence on KITTI and fine-tuned for the image semantic segmentation task. mIOU is used for evaluation.}
  \label{tab:kitti_semseg}
  
\end{table}

\begin{figure*}
\centering
\begin{overpic}[width=1\textwidth]{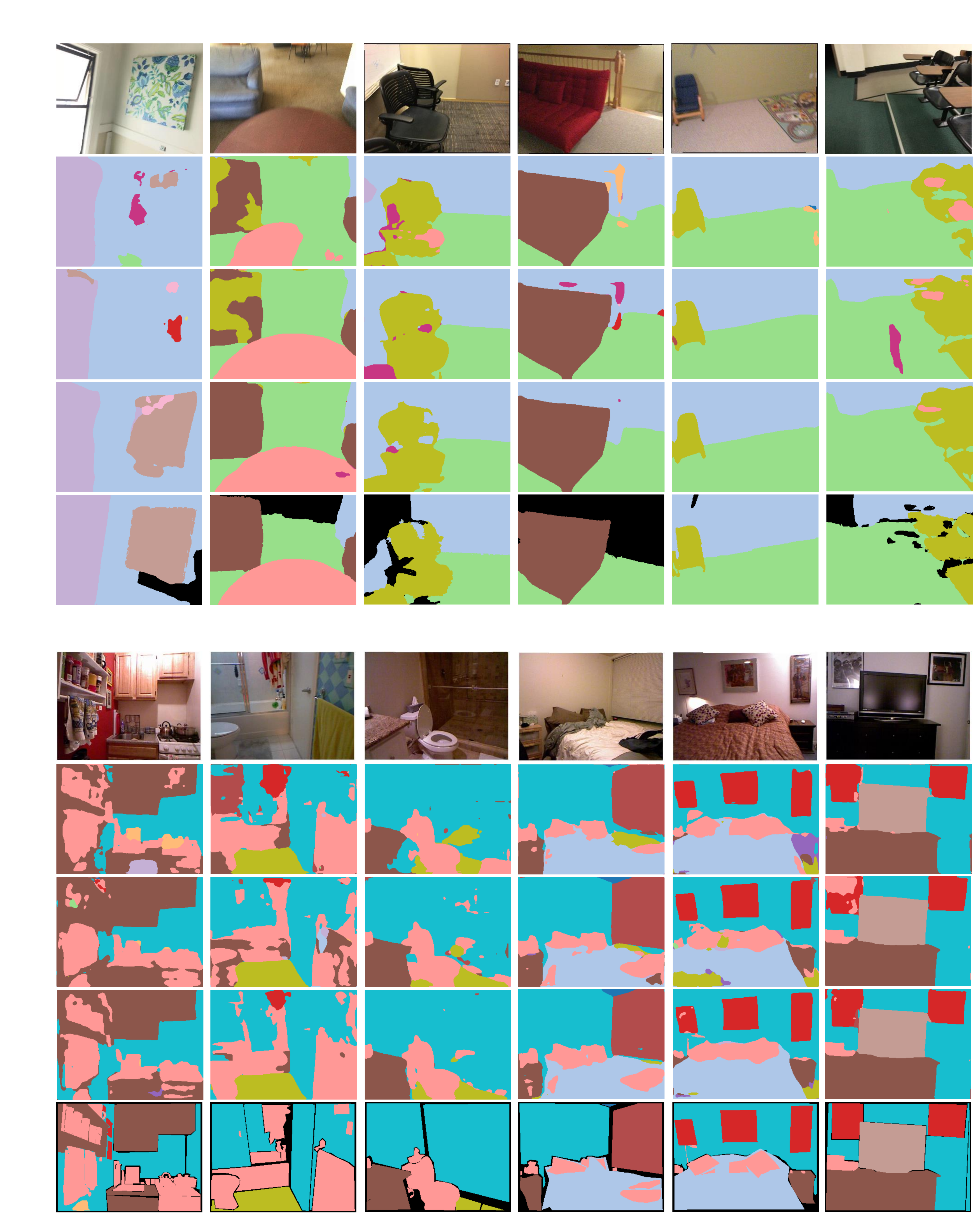}{
 		\put(39,97.3) {\textbf{ScanNet}}
 		\put(38.5,47.7){ \textbf{NYUv2}}
 		
 		\put(-0.5,92){\small {Input}} 
 		\put(-2.50,82){\small {ImageNet}}
 		\put(-1,73){\small {Pri3D}}
 		\put(0,64){\small {Ours}}
 		\put(0.5,55){\small {GT}}

 		\put(-0.5,42){\small {Input}} 
 		\put(-2.5,32.5){\small{ImageNet}}
 		\put(-1,23.5){\small {Pri3D}}
 		\put(0,14.5){\small {Ours}}
 		\put(0.5,5){\small {GT}}
    }\end{overpic}

  \caption{\textbf{More qualitative results of semantic segmentation task on ScanNet and NYUv2 datasets.} All the methods are pre-trained on ScanNet with ResNet50 as the backbone encoder. 
  }
  \label{fig:vis_semseg2d_sup}
    
\end{figure*}

 \section{Performance on KITTI Dataset}
 
We further test our method on KITTI \cite{geiger2013vision} dataset, to see the effectiveness of our method in the outdoor autonomous driving scenario. KITTI is a dataset captured by driving around a city with cars equipped with different kinds of sensors, including stereo camera, GPS, laser-scanner, etc. In our experiment, we use the unlabeled RGB-D sequences for pre-training and fine-tune the pre-trained network on the 2D image semantic segmentation task.  To compute the geometric consistency sets, we use the Voxel Cloud Connectivity Segmentation (VCCS) \cite{papon2013voxel} method implemented by PCL \cite{Rusu_ICRA2011_PCL} library to extract clusters on the per-view point clouds. During training, for each view pair, we use the geometric consistency sets from one view and project them onto the other, while the geometric consistency sets from the other view are ignored. In this way, we can have the same geometric consistency sets for the corresponding views. Table \ref{tab:kitti_semseg} illustrates the results. We note that the point clouds in KITTI are partial and noisy, and the moving objects in the scenes may lead to incorrect correspondences between views, which makes the results sensitive to different clustering parameters. The experiment here is to demonstrate the possibility of adapting our method to outdoor scenes; we hope our work can motivate future research in this direction. 
 

\section{More Qualitative Results}

In Figure \ref{fig:vis_semseg2d_sup}, we show more qualitative results of 2D semantic segmentation task on ScanNet\cite{dai2017scannet} and NYUv2\cite{silberman2012indoor} datasets. As is shown, the segmentation results produced with our method have less noise compared with those from other methods.

\end{appendices}

\end{document}